\documentclass[10pt,journal,compsoc]{IEEEtran}
\ifCLASSOPTIONcompsoc
  \usepackage[nocompress]{cite}
\else
  \usepackage{cite}
\fi

\ifCLASSINFOpdf
\else
\fi
\usepackage{times}
\usepackage{epsfig}
\usepackage{graphicx}
\usepackage{amsmath}
\usepackage{amssymb}

\usepackage{graphicx}
\usepackage{amsmath}
\usepackage{amsthm}
\usepackage{amsfonts} 
\usepackage{mathabx}
\usepackage{bm}
\usepackage{subcaption}
\usepackage{color}
\usepackage{multirow}
\usepackage{stackengine}
\usepackage[switch]{lineno}
\usepackage{multirow}
\usepackage{color}
\usepackage{booktabs}

\usepackage{array}
\usepackage{colortbl}

\usepackage{algorithm}
\usepackage{algorithmic}

\usepackage{makecell}
\usepackage{soul}
\usepackage{pifont}

\newcommand{\bmmc}[1]{\bm{\mathcal{#1}}}

\usepackage{tikz}
\newcommand{\ballnumber}[1]{\tikz[baseline=(myanchor.base)] \node[circle,draw=black,fill=white,inner sep=1pt] (myanchor) {\color{black}\bfseries\footnotesize #1};}

\begin{document}
%
\title{Algorithm and Hardware Co-Design of Energy-Efficient LSTM Networks for Video Recognition with Hierarchical Tucker Tensor Decomposition}
%
%
%
%

\author{Yu Gong, 
        Miao Yin, 
        Lingyi Huang, 
        Chunhua Deng,
        Yang Sui,
        and Bo Yuan
\IEEEcompsocitemizethanks{\IEEEcompsocthanksitem Yu Gong, Miao Yin, Lingyi Huang, Yang Sui and Bo Yuan are with the Department
of Electrical and Computer Engineering, Rutgers University, Piscataway, NJ, 08854.
E-mail: yg430@soe.rutgers.edu, miao.yin@rutgers.edu, lingyi.huang@rutgers.edu, yang.sui@rutgers.edu, bo.yuan@soe.rutgers.edu.
\IEEEcompsocthanksitem Chunhua Deng is currently with ScaleFlux Inc. This work was done when the author was with Rutgers University. E-mail: chunhua.deng518@gmail.com. 
\IEEEcompsocthanksitem Yang Sui contributed to this work but was accidentally omitted in the published version.
}
}

\IEEEtitleabstractindextext{%
\begin{abstract}

Long short-term memory (LSTM) is a type of powerful deep neural network that has been widely used in many sequence analysis and modeling applications. However, the large model size problem of LSTM networks make their practical deployment still very challenging, especially for the video recognition tasks that require high-dimensional input data. Aiming to overcome this limitation and fully unlock the potentials of LSTM models, in this paper we propose to perform algorithm and hardware co-design towards high-performance energy-efficient LSTM networks. At algorithm level, we propose to develop \textit{fully decomposed hierarchical Tucker (FDHT)} structure-based LSTM, namely FDHT-LSTM, which enjoys ultra-low model complexity while still achieving high accuracy. In order to fully reap such attractive algorithmic benefit, we further develop the corresponding customized hardware architecture to support the efficient execution of the proposed FDHT-LSTM model. With the delicate design of memory access scheme, the complicated matrix transformation can be efficiently supported by the underlying hardware without any access conflict in an on-the-fly way. Our evaluation results show that both the proposed ultra-compact FDHT-LSTM models and the corresponding hardware accelerator achieve very high performance. Compared with the state-of-the-art compressed LSTM models, FDHT-LSTM enjoys both order-of-magnitude reduction (more than $1000 \times$) in model size and significant accuracy improvement (0.6\% to 12.7\%) across different video recognition datasets. Meanwhile, compared with the state-of-the-art tensor decomposed model-oriented hardware TIE, our proposed FDHT-LSTM architecture achieve $2.5\times$, $1.46\times$ and $2.41\times$ increase in throughput, area efficiency and energy efficiency, respectively on LSTM-Youtube workload. For LSTM-UCF workload, our proposed design also outperforms TIE with $1.9\times$ higher throughput, $1.83\times $ higher energy efficiency and comparable area efficiency.
    
\end{abstract}

}

\maketitle

\IEEEdisplaynontitleabstractindextext

%
\IEEEpeerreviewmaketitle

\ifCLASSOPTIONcompsoc
\IEEEraisesectionheading{\section{Introduction}\label{sec:introduction}}
\else
\section{Introduction}
\label{sec:introduction}
\fi

%
%
%
%
\IEEEPARstart{L}{ong} short-term memory (LSTM) networks are popularly used in many real-world sequential analysis and processing tasks. Thanks
to their inherent strong capability of capturing the
temporal dependency and modeling the sequential correlation,
the state-of-the-art LSTMs have achieved unprecedented success in many
important temporal sequence-involved applications, such as video recognition \cite{donahue_long-term_nodate}, speech recognition \cite{graves_hybrid_2013} and natural language processing (NLP) \cite{azari_energy-efficient_2019}.

Despite their current widespread adoptions, the \textit{large model sizes} of LSTM networks make the practical deployment still very challenging. To be specific, because the input data of many real-world applications, e.g., video processing and NLP, naturally exhibit high dimensionality, the corresponding input-to-hidden weight matrices of LSTMs are typically extremely large. For instance, as analyzed in \cite{yang2017tensor}, even with small-size (256 states) hidden layer, the LSTM evaluated on UCF11 video recognition dataset \cite{liu2009recognizing} already requires more than 50 million weights. Evidently, such ultra-large model size poses a series of severe challenges for the efficient deployment of LSTMs, including but not limited to high memory footprint, long processing time, low energy efficiency, and insufficient accuracy incurred by high difficulty of training. 

To address these challenges, various model compression methods, such as pruning and quantization, have been proposed to build compact LSTMs. Consider the compression ratio provided by these methods are typically limited and may be insufficient for the very large LSTMs, \textit{tensor decomposition}, as a low-rank approximation approach, have attracted a lot of attention towards high-performance compression of LSTM models. In general, tensor decomposition aims to factorize a large-size tensor to a series of small tensor cores with maintaining small approximation error. After performing tensor decomposition, a very elegant mathematical phenomenon is that the resulting space and time complexity can be significantly reduced, sometimes even with order-of-magnitude reduction. Motivated by this attractive property, recently several different tensor decomposition approaches, such as tensor train (TT), tensor ring (TR) and block-term (BT), have been adopted to build compact LSTMs models \cite{deng_tie_2019} \cite{pan2019compressing} \cite{ye2018learning} for various temporal data analysis tasks, e.g., video recognition and long-term dynamic forecasting. Compared with the original uncompressed large-size LSTMs, these tensor decomposed models exhibit much smaller memory footprints and competitive classification/prediction accuracy.

Although the state-of-the-art tensor decomposed LSTMs demonstrate their promising potentials, they are still facing two inherent limitations. \ul{First}, the existing approaches can only factorize the input-to-hidden layers instead of the entire model, thereby limiting the overall compression performance. Notice that, as will be shown in Section \ref{sec:alg}, the straightforward decomposition on the hidden-to-hidden layers cannot solve this problem due to the significant accuracy degradation. \ul{Second}, the tensor decomposition approaches adopted in the existing LSTM compression works have inherent limitations. To be specific, TT decomposition requires that the border tensor cores be rank-1 tensors, thereby directly limiting the overall representation power. Also, using BT decomposition brings extra flatten and permutation operations, which causes significant computation overhead for the resulting BT-LSTM models. More importantly, as will be analyzed in Section \ref{sec:alg}, the existing adopted tensor decomposition approaches (TT, TR and BT) in model compression, theoretically, do not provide the best space complexity reduction. Consequently, they are not the ideal solutions for building tensor factorized high-accuracy ultra-compact LSTM models.

\begin{figure}[t]
    \centering
    \includegraphics[width=0.95\linewidth]{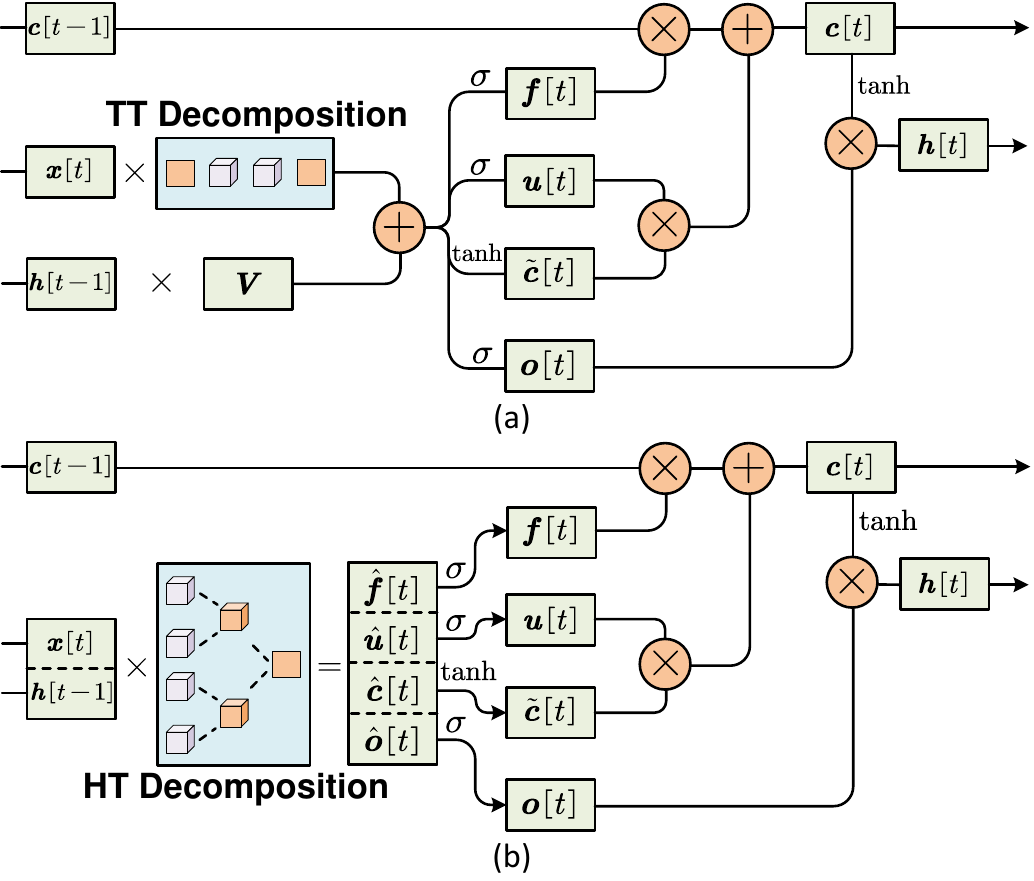}
    \caption{Architecture of tensor decomposition-based LSTM. (a) Prior TT-LSTM. (b) The proposed FDHT-LSTM. Reproduced from \cite{yin2021towards}.}
    \label{fig:cht_lstm}
\end{figure}

To overcome these limitations and develop high-performance LSTM models, in this paper \footnote{This paper is the extension of the authors' prior work \cite{yin2021towards}.} we propose to leverage \textit{Hierarchical Tucker (HT)} \cite{hackbusch2009new} technique, an under-explored yet powerful tensor tool, to \textit{fully decompose} the LSTM networks. The resulting compressed model, namely \textit{FDHT-LSTM}, enjoys ultra-low complexity with still maintaining high accuracy. Then, to fully reap the benefits brought by the proposed efficient compression approach, we further develop the corresponding hardware architecture to accelerate the execution of FDHT-LSTM models. Overall, the contributions of this paper are summarized as follows:

\begin{itemize}
    \item At the algorithm level, we propose to impose fully decomposed hierarchical Tucker (FDHT) structure on the LSTM networks to achieve ultra-high compression performance. The proposed FDHT structure contains two main features (as shown in Figure \ref{fig:cht_lstm}). \ul{First}, the underlying LSTM is built via using HT decomposition, a powerful approach that can properly capture and model the correlation and structure in high-dimensional data. \ul{Second}, the entire LSTM model, instead of one or few component layers, is fully decomposed in the HT format in a homogeneous way. In other words, the low-rank correlation among different component layers are fully exploited, and thereby bringing order-of-magnitude reduction in model size while still achieving very high accuracy.

    \item At the hardware level, we design and implement the corresponding FDHT-LSTM hardware architecture to fully leverage the algorithmic benefits. In order to support various complicated matrix transformations introduced by the computation flow of FDHT-LSTM, we first propose to utilize 2-D SRAM array to properly map the high-order tensor transpose on the physical 2-D memory component. Based on this philosophy, we further propose novel write and read access schemes to the 2-D SRAM array, thereby enabling conflict-free memory access with high flexibility and reconfigurability to different workloads.  
    
    \item Evaluation results demonstrate the superior performance of our proposed FDHT-LSTM and the corresponding hardware accelerator. Experiments show that the FDHT-LSTM models can use very few parameters (less than 10,000 weights) to achieve very high accuracy across different video recognition datasets. Compared with the state-of-the-art compressed LSTM models, FDHT-LSTM enjoys both order-of-magnitude reduction (more than $1000 \times$) in model size and significant accuracy improvement (0.6\% to 12.7\%). Meanwhile, compared with the state-of-the-art tensor decomposed model-oriented hardware TIE, our proposed FDHT-LSTM architecture achieves $2.5\times$, $1.46\times$ and $2.41\times$ increase in throughput, area efficiency and $2.41\times$ energy efficiency, respectively on LSTM-Youtube workload. For LSTM-UCF workload, our proposed design also outperforms TIE with $1.9\times$ higher throughput, $1.83\times $ higher energy efficiency and comparable area efficiency.

\end{itemize}

The rest of this paper is organized as follows. Section \ref{sec:background} introduces the background of tensor decomposition and the motivation of hierarchical Tucker (HT)-based LSTM compression. The proposed fully decomposed HT-structured (FDHT) LSTM algorithm is described in Section \ref{sec:alg}. Section \ref{sec:hw} presents the corresponding FDHT-LSTM hardware architecture. The evaluation performance at both algorithm and hardware levels is reported in Section \ref{sec:eval}. Section \ref{sec:conclu} draws the conclusions. 

\section{Background and Motivation}
\label{sec:background}

\subsection{Preliminaries}
\label{subsec:prelim}

\subsubsection{Notation}

In this paper, boldface lower-case, boldface capital and boldface calligraphic letters represent vectors, matrices and higher-order tensors, respectively, e.g. $\bm{x}$, $\bm{X}$ and $\bm{\mathcal{X}}$. Additionally, letters with indices in bracket denote the entry, e.g., $\bm{x}(i)$, $\bm{X}(i,j)$, $\bm{\mathcal{X}}(i_1,\cdots,i_d)$.

 \subsubsection{Tensor Contraction} 
Tensor contraction is the multiplication between two higher-order tensors where more than one dimension matches. For example, given two tensors $\bm{\mathcal{A}}\in\mathbb{R}^{M_1\times M_2 \times C}$ and $\bm{\mathcal{B}}\in\mathbb{R}^{C\times N_1\times N_2}$, where the 3rd dimension of $\bm{\mathcal{A}}$ is identical to the 1st dimension of $\bm{\mathcal{B}}$ with size $C$, the result of tensor contraction $\bmmc{C}=\bmmc{A}\times_{1}^{3}\bmmc{B}$, as a size- $M_1\times M_{2}\times N_1 \times N_2$ tensor, can be calculated as:
\begin{equation}
\label{eqn:contraction}
    \bmmc{C}(i_1, i_2, j_1, j_2)=\sum_{k=1}^{C}\bm{\mathcal{A}}(i_1, i_2, k)\bm{\mathcal{B}}(k, j_1, j_2).
\end{equation}

\begin{figure}[t!]
    \centering
    \includegraphics[width=0.95\linewidth]{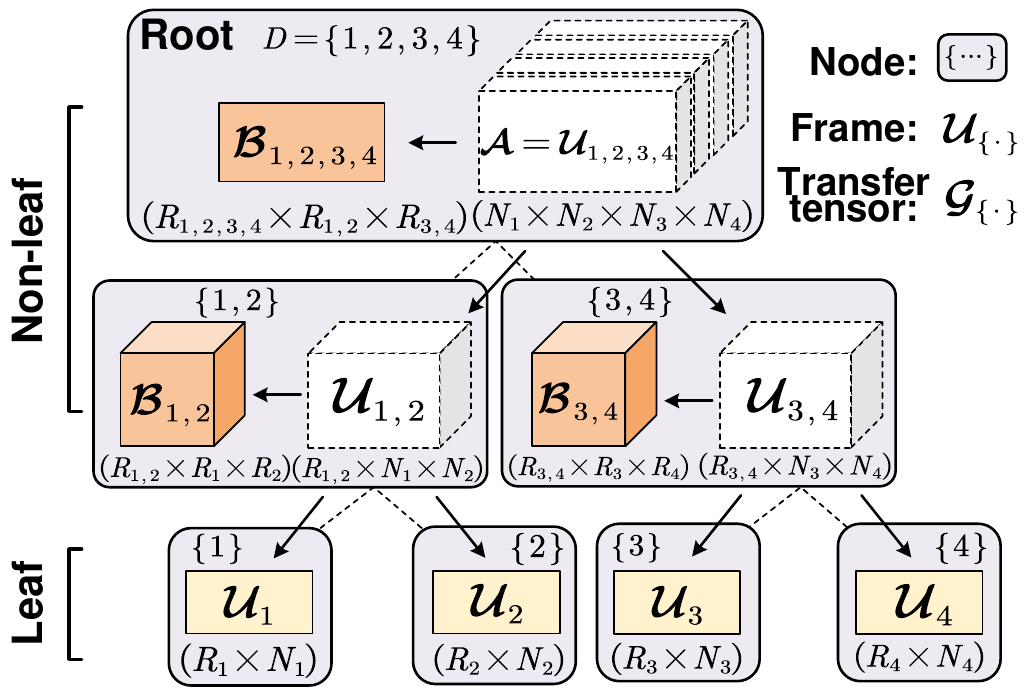}
    \caption{Standard HT decomposition example for a 4-order tensor. The hierarchy is a binary tree with root $\{1,2,3,4\}$, where nodes are denoted by rounded rectangles. $\{1\}$, $\{2\}$, $\{3\}$ $\{4\}$ are leaf nodes, and $\{1,2\}$, $\{3,4\}$ are their parents. In HT format, only colored leaf frames and transfer tensors are needed to store. Reproduced from \cite{yin2021towards}.}
    \label{fig:htd_example}
    \vspace{-2mm}
\end{figure}

\subsubsection{Hierarchical Tucker Decomposition} 
As a special case of tensor decomposition, hierarchical Tucker (HT) approach recursively decomposes the original tensor into small tensor cores with hierarchical levels from top to bottom in a binary tree. As illustrated in Figure \ref{fig:htd_example}, the upper-level intermediate components can be factorized to lower-level components. Here the intermediate components are referred as \textit{frames}, each of which corresponds to a unique \textit{node}. Additionally, each node in the binary tree is represented by a \textit{dimension set}. In general, given a $d$-order tensor $\bm{\mathcal{A}}\in\mathbb{R}^{N_1\times\cdots\times N_d}$, we can build a binary tree to split the original dimension set $\{1,\cdots,d\}$ and assign the index to each node. To be specific, the root node $D=\{1,2,\cdots,d\}$ is associated with the root frame $\bm{\mathcal{A}}=\bm{\mathcal{U}}_D$, $s\subsetneq D$ is the node corresponding to frame $\bm{\mathcal{U}}_s$, and $s_1, s_2\subsetneq s$ are the left and right child nodes of node $s$. Notice that if we define $\mu_s=\min(s), \nu_s=\max(s)$, then each non-leaf frame $\bm{\mathcal{U}}_{s}\in\mathbb{R}^{R_{s}\times N_{\mu_s}\times\cdots\times N_{\nu_s}}$ can be recursively decomposed to frames ($\bm{\mathcal{U}}_{s_1}$ and $\bm{\mathcal{U}}_{s_2}$) and transfer tensor $\bm{\mathcal{B}}_s\in\mathbb{R}^{R_s\times R_{s_1} \times R_{s_2}}$ as

\begin{equation}
\bm{\mathcal{U}}_s=\bm{\mathcal{B}}_s\times_{1}^{2}\bm{\mathcal{U}}_{s_1}\times_{1}^{2}\bm{\mathcal{U}}_{s_2},
\label{eqn:htd}
\end{equation}
where $R_s, R_{s_1}, R_{s_2}$ are referred as \textit{hierarchical ranks}. Overall, by using hierarchical Tucker decomposition that recursively decomposes the frames from top to bottom, we can use the combination of the small-size 2-order leaf frames and 3-order transfer tensors to store the original large-size $d$-order tensor $\bm{\mathcal{X}}=\bm{\mathcal{U}}_D$.

\subsection{Compressing LSTMs via Tensor Decomposition}

In the recent years tensor decomposition has emerged as a very attractive technique to compress very large LSTM models. In \cite{yang2017tensor}, tensor train (TT) decomposition is applied to factorize the input-to-hidden layer of LSTMs and GRUs for video recognition. The reported experimental results show that very significant compression ratio can be achieved while maintaining high accuracy. Motivated by this success, other advanced tensor decomposition approach, such as block-term (BT) \cite{ye2018learning} decomposition and tensor ring (TR) decomposition \cite{ye2018learning}, are also used to build compact LSTM models (BT-LSTM  and TR-LSTM) in video processing tasks. In addition, tensor decomposed LSTMs also demonstrate high performance in other sequential analysis and prediction tasks. For instance, \cite{yu2017long} proposes a TT-format LSTM to perform long-term forecasting in dynamic systems. 

\subsection{Motivation of Compressing LSTM via HT Decomposition}

In this paper we propose to adopt HT decomposition, a relatively little noticed yet powerful tensor factorization approach for LSTM compression. Our choice is motivated by two reasons. First, compared with its well-explored counterparts (e.g. TT, TR and BT decomposition) in the model compression field, HT decomposition, by it nature, can provide higher space complexity reduction on the same-size tensor data with the same selected rank, thereby implying that even smaller LSTM models can be constructed in the HT format. Second, from the perspective of tensor theory, the inherent hierarchical structure of HT also enables better weight sharing and hierarchical representation from high-dimensional data, which are very important to guarantee the representation capability of tensor decomposed LSTM models. As we will report in Section \ref{sec:eval}, the HT structure-based LSTM models consistently demonstrate superior  performance than the existing compressed LSTMs using other tensor decomposition methods with respect to both model accuracy and compression ratio.

\section{The Proposed FDHT-LSTM: Algorithm}
\label{sec:alg}

\subsection{Compact HT-structure Linear Layer}
\label{sec:ht-layer}

Consider the linear layer is the foundation of LSTM models, in this subsection we first build a compact HT-structure layer that serves as the key component of the proposed FDHT-LSTM model.

\subsubsection{Tensorization} 
In the compact HT-layer, computations are performed with high-order tensor contraction, thus we first need to transform all the original variables into tensor format. In general, for an original linear layer with weight matrix $\bm{W}\in\mathbb{R}^{O\times I}$ that linearly maps an input vector $\bm{x}\in\mathbb{R}^{I}$ to an output vector $\bm{y}\in\mathbb{R}^{O}$ in HT-based format, we first transform the weight matrix $\bm{W}$ to an $d$-order tensor $\bm{\mathcal{W}}\in\mathbb{R}^{O_1\times\cdots\times O_d\times I_1\times\cdots\times I_d}$. In addition, by defining $I=\prod_{i=1}^{d}I_i$ and $O=\prod_{j=1}^{d}O_j$, we can also tensorize the affiliated input vector $\bm{x}$ and output vector $\bm{y}$ to input and output tensor $\bm{\mathcal{X}}\in\mathbb{R}^{I_1\times\cdots\times I_d}$ and $\bm{\mathcal{Y}}\in\mathbb{R}^{O_1\times\cdots\times O_d}$, respectively. 


\subsubsection{Decomposing  $\mathcal{W}$} 
After tensorizing the large-size weight matrix $\bm{W}$ as weight tensor $\bm{\mathcal{W}}\in\mathbb{R}^{O_1\times\cdots\times O_d\times I_1\times\cdots\times I_d}$, we can then utilize HT decomposition to denote the original $\bm{W}$ with a set of small-size matrices and tensors. With the definition of HT decomposition in Eq. (\ref{eqn:htd}), $\bm{\mathcal{W}}$ can be factorized as
\begin{equation}
\begin{aligned}
\bm{\mathcal{W}}(o_1,\cdots,o_d,i_1,\cdots,i_d)=
\sum_{k=1}^{R_D}\sum_{p=1}^{R_{D_1}}\sum_{q=1}^{R_{D_2}}\bm{\mathcal{B}}_D(k,p,q)\cdot\\
\bm{\mathcal{U}}_{D_1}(p,\varphi_{D_1}(\bm{o},\bm{i}))\bm{\mathcal{U}}_{D_2}(q,\varphi_{D_2}(\bm{o},\bm{i})),
\label{eqn:w_ht}
\end{aligned}
\end{equation}
\noindent where $\varphi_s(\bm{o},\bm{i})$ is a mapping function which returns the associate indices $\bm{o}=(o_1,\cdots,o_d)$ and $\bm{i}=(i_1,\cdots, i_d)$ for a specified frame $\bm{\mathcal{U}}_s$ with a given node $s$ and $d$. For example, with $d=6$ and $s=\{3,4\}$, the output of $\varphi_s(\bm{o},\bm{i})$ is $(o_3,o_4,i_3,i_4)$. Additionally, $\bm{\mathcal{U}}_{D_1}$ and $\bm{\mathcal{U}}_{D_2}$ can be further recursively decomposed as
\begin{equation}
\begin{aligned}
(\bm{\mathcal{U}}_s)(k,\varphi_s(\bm{o},\bm{i}))
=\sum_{p=1}^{R_{s_1}}\sum_{q=1}^{R_{s_2}}(\bm{\mathcal{B}}_{s})(k,p,q)\\
\cdot(\bm{\mathcal{U}}_{s_1})(p,\varphi_{s_1}(\bm{o},\bm{i}))(\bm{\mathcal{U}}_{s_2})(q,\varphi_{s_1}(\bm{o},\bm{i})),
\end{aligned}
\end{equation}
\noindent where $D=\{1,2,\cdots,d\}$, $D_1=\{1,\cdots,\lfloor d/2\rfloor\}$ and $D_2=\{\lceil d/2\rceil,\cdots,d\}$ are the left and right child nodes of the root node $D$.


\subsubsection{HT-Structure Layer} 
As the original weight matrix is decomposed to HT-format, the HT-structured linear layer can be readily constructed via HT-format matrix-vector multiplication instead of recovering back to original format. To be specific, the forward pass of an HT-layer is computed as:
\begin{equation}
\begin{aligned}
\bm{\mathcal{Y}}(\bm{o})=\sum_{\bm{i}}
\sum_{k=1}^{R_D}\sum_{p=1}^{R_{D_1}}\sum_{q=1}^{R_{D_2}}(\bm{\mathcal{B}}_D)(k,p,q)\cdot\\
\bm{\mathcal{U}}_{D_1}(p,\varphi_{D_1}(\bm{o},\bm{i}))\bm{\mathcal{U}}_{D_2}(q,\varphi_{D_2}(\bm{o},\bm{i}))\bm{\mathcal{X}}(\bm{i}).
\end{aligned}
\end{equation}
After the above computation, we need to reshape the output tensor $\bmmc{Y}$ back to the original vector-format $\bm{y}$. Hence, we can simplify the entire computing process as a single HT-structure layer:
\begin{equation}
\bm{y}= HTL(\bm{W}, \bm{x}).
\end{equation}

\begin{figure}[t!]
    \centering
    \includegraphics[width=\linewidth]{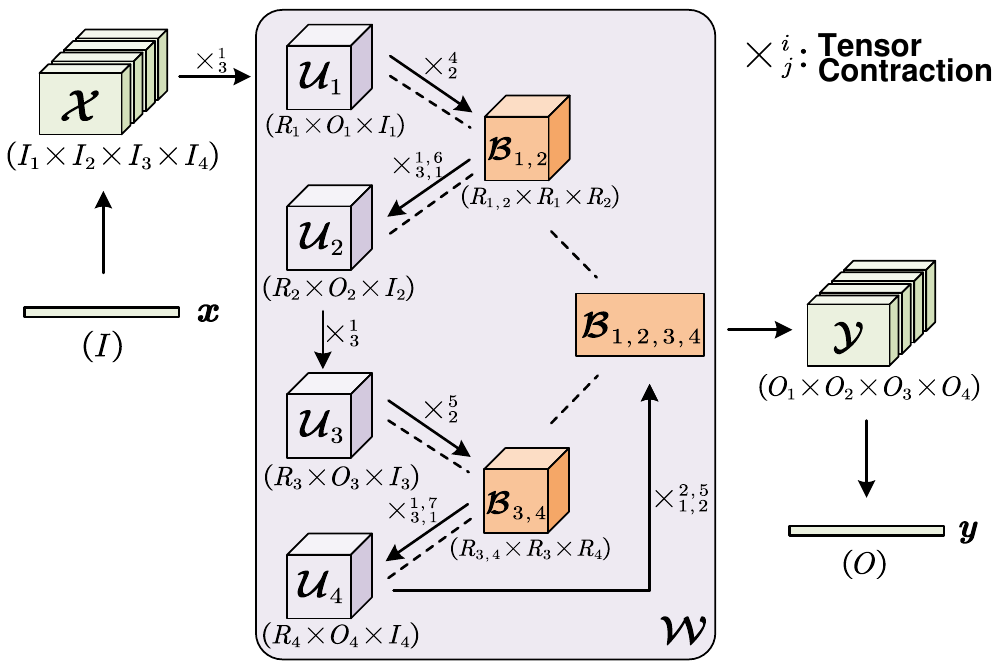}
    \caption{Computation flow in the HT-layer for $d$=4. Arrows represent the directions of tensor contractions among the decomposed tensors. Here leaf frames $\{\bm{\mathcal{U}}\}_{i=1}^{4}$ are 3-order tensors since the weight tensor is tensorized from a matrix. Reproduced from \cite{yin2021towards}.}
    \label{fig:htl}
\end{figure}

As depicted in Figure \ref{fig:htl}, the consecutive arrows denote the computation flow in the computation process of an HT-structure layer. Specifically, the input vector $\bm{x}$ is first tensorized, and then it is contracted with the HT-format weight tensor in a hierarchical way. Finally, the obtained output tensor is reshaped back to output vector $\bm{y}$.

\subsubsection{Benefits on Low Cost} 
\begin{figure}[t]

    \centering
    \includegraphics[width=\linewidth]{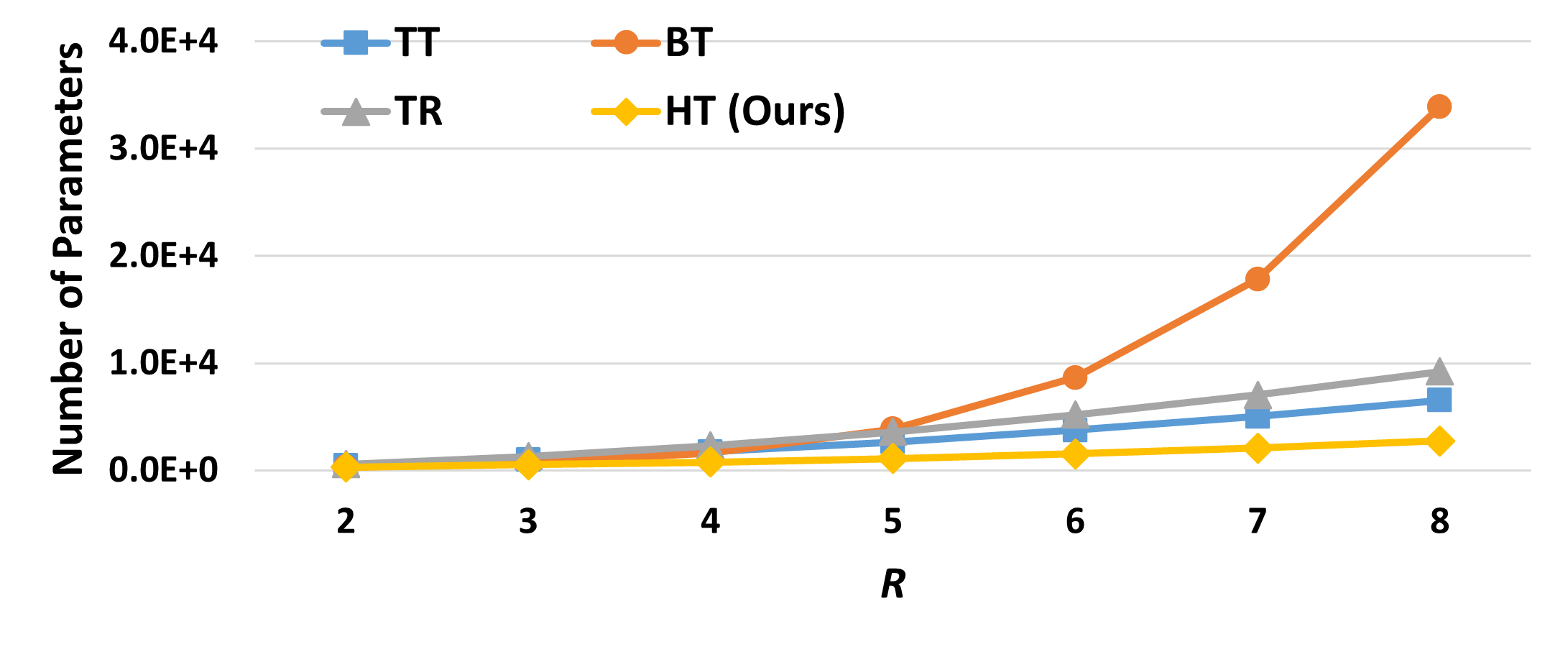}
    \vspace{-4mm}
    \caption{Curves for number of parameters in different tensor-format layer with respect to tensor rank $R$. Here we use settings in the related works, i.e., $d=5$, $(I_1,\cdots,I_5)=(8,10,10,9,8)$, $(O_1,\cdots,O_5)=(4,4,2,4,2)$. Reproduced from \cite{yin2021towards}.}
    \label{fig:complexity}

\end{figure}

With the proposed HT-layer, one most important benefit is huge complexity reduction. Table \ref{tab:complexity} shows the theoretical space complexity in comparison with other tensor factorized linear layer as well as the uncompressed one. Considering tensor rank $R$ is generally smaller than $I$ or $O$, HT-structure layer can provide the lowest space complexity as shown in this table. Additionally, to verity the practical ability of parameters reduction, we plot curves of number of parameters with respect to $R$ in different tensor decomposition formats. As shown in Figure \ref{fig:complexity}, to store size-$57,600 \times256$ weight matrix used in \cite{yang2017tensor,ye2018learning,pan2019compressing}, our HT-structure layer requires the fewest number of parameters with the same rank settings among all different types of tensor factorization methods. As will be shown in Section \ref{sec:eval}, such advantages will further be verified via empirical experimental results across various popular video recognition datasets.

\begin{table}
\normalsize
\caption{Space complexities for different tensor-format linear layers. Here $R=\max_{s\subsetneq D}R_s$, $I'=\max_{k\in D}I_{k}$, $O'=\max_{k\in D}O_{k}$. Reproduced from \cite{yin2021towards}.}
\def\arraystretch{1.15}
\centering
\begin{tabular}{c|c} 
\hline

\textbf{Compressed Linear Layer}  & \multicolumn{1}{c}{\textbf{Space Complexity}}  \\
\hline\hline
Uncompressed &  \multirow{1}{*}{$\mathcal{O}(I'O')$} \\
\hline
Tensor Train (TT)-structure &  \multirow{1}{*}{$\mathcal{O}(dI'O'R^2)$} \\
\hline
Tensor Ring (TR)-structure &  \multirow{1}{*}{$\mathcal{O}(dI'O'R^2)$} \\
\hline
Block-Term (BT)-structure &  \multirow{1}{*}{$\mathcal{O}(dI'O'R+R^d)$} \\
\hline
Hierarchical Tucker (HT)-structure & \multirow{1}{*}{$\mathcal{O}(dI'O'R+dR^3)$} \\

\hline
\end{tabular}
\label{tab:complexity}

\end{table}

\subsection{Fully Decomposing the HT-structure LSTM}

\subsubsection{Challenges of Fully Compressing LSTM}
\begin{figure}[t]
    \centering
    \includegraphics[width=\linewidth]{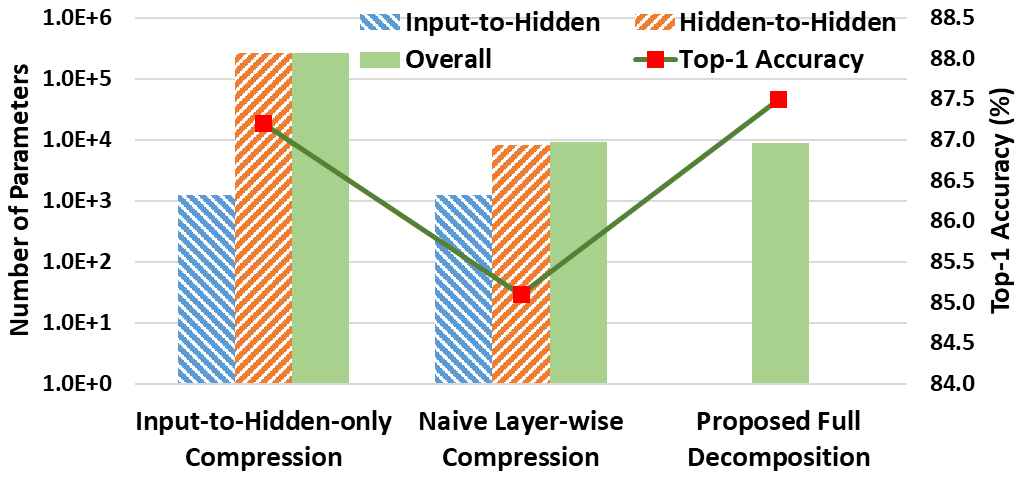}
    \caption{Comparison between naive layer-wise compression and our proposed full decomposition in terms of number of parameters and test accuracy on UCF11 dataset. Reproduced from \cite{yin2021towards}.}
    \label{fig:full_compress}

\end{figure}

Recall that a typical LSTM model consists of  input-to-hidden layer and hidden-to-hidden layer. Built on the top of the underlying HT structure for each component layer, performing fully decomposition on the entire LSTM models can evidently bring further performance improvement. To that end, one naive way is to simply compress each layer with HT decomposition. In other words, all weight matrices in the LSTM model are independently decomposed to HT format. Although this strategy can indeed bring full compression on the entire LSTM network, such straightforward layer-wise compression strategy suffers from huge accuracy drop. Figure \ref{fig:full_compress} shows the experimental results using the layer-wise compression with HT decomposition. Here more than $2\%$ accuracy drop can be observed as compared to the conventional input-to-hidden-only compression adopted in \cite{yang2017tensor,ye2018learning,pan2019compressing}. Overall, fully compressing the entire LSTM model using HT decomposition without accuracy drop is challenging and non-trivial.

\subsubsection{Proposed FDHT-LSTM}
To fully compress LSTM models without performance degradation, we propose a homogeneous compression method, namely \textit{fully decomposed HT (FDHT)} decomposition, to obtain highly-compact LSTM models. Figure \ref{fig:cht_lstm} illustrates the key idea of the proposed full decomposition. To be specific, in order to maximally leverage the linear correlation across all weight matrices in the entire LSTM model, we first concatenate all the weight matrices as a single huge matrix such that the entire model can be considered as a ``mega'' linear layer. Then, at each time step, all the intermediate results only need one-time multiplication in the forward propagation process:
\begin{equation}
\begin{aligned}
\begin{bmatrix}
\hat{\bm{f}}[t]\\
\hat{\bm{u}}[t]\\
\hat{\bm{c}}[t]\\
\hat{\bm{o}}[t]
\end{bmatrix}&=
\begin{bmatrix}
\bm{W}_f & \bm{V}_f\\
\bm{W}_u & \bm{V}_u\\
\bm{W}_c & \bm{V}_c\\
\bm{W}_o & \bm{V}_o
\end{bmatrix}
\begin{bmatrix}
\bm{x}[t]\\
\bm{h}[t-1]
\end{bmatrix}\\
&=\bm{W}\bm{I}[t].
\end{aligned}
\label{eqn:entirety}
\end{equation}

With the above interpretation, we can directly impose the desired HT structure on the integrated single huge matrix in an LSTM model. To be specific, following the scheme described in Section \ref{sec:ht-layer}, we can tensorize and decompose the entire LSTM models to the HT format, and then perform forward pass as:
\begin{equation}
    \begin{bmatrix}
\hat{\bm{f}}[t]\\
\hat{\bm{u}}[t]\\
\hat{\bm{c}}[t]\\
\hat{\bm{o}}[t]
\end{bmatrix} = \bm{Z}[t]=HTL(\bm{W}, \bm{I}[t]).
\end{equation}
\noindent Correspondingly, the outputs of the FDHT-LSTM can be calculated as follows:
\begin{equation}
\begin{aligned}
\bm{f}[t]=&\sigma(\hat{\bm{f}}[t]+\bm{b}_f)\\
\bm{u}[t]=&\sigma(\hat{\bm{u}}[t]+\bm{b}_u)\\
\bm{c}[t]=&\bm{f}[t]\odot\bm{c}[t-1]+\bm{u}[t]\odot\tanh(\hat{\bm{c}}[t]+\bm{b}_c)\\
\bm{o}[t]=&\sigma(\hat{\bm{o}}[t]+\bm{b}_o)\\
\bm{h}[t]=&\bm{o}[t]\odot\tanh(\bm{c}[t]),
\end{aligned}
\end{equation}
\noindent where $\sigma$, $\tanh$ and $\odot$ are the sigmoid function, hyperbolic function and element-wise product, respectively.

\subsubsection{HT-based Gradient Calculation}
Notice that in order to ensure that the desired FDHT-LSTM model can be properly obtained, the backward propagation in the training process should also be reformulated to HT-based format. In general, given an HT-structure linear layer $\bm{\mathcal{W}}=\bm{\mathcal{U}}_D$ and $\frac{\partial\bm{\mathcal{Y}}}{\partial\bm{\mathcal{U}}_D}=\bm{\mathcal{X}}$, and defining $s$, $F(s)$ and $B(s)$ as non-root node, parent and sibling nodes of node $s$ in the binary tree, respectively, the partial derivative of the output tensor with respect to frames can be recursively calculated as:
\begin{equation}
\begin{aligned}
\frac{\partial\bm{\mathcal{Y}}}{\partial\bm{\mathcal{U}}_s}=&
\bm{\mathcal{B}}_{F(s)}\times_{1}^{3}\bm{\mathcal{U}}_{B(s)}\\
&\times_{\substack{1,\cdots,\nu_{B(s)}-\mu_{B(s)}+2,\nu_{F(s)}-\mu_{F(s)}+3,\\\cdots,\nu_{F(s)}-\mu_{F(s)}+\nu_{B(s)}-\mu_{B(s)}+3}}^{1,3,\cdots,2\nu_{B(s)}-2\mu_{B(s)}+4}\frac{\partial\bm{\mathcal{Y}}}{\partial\bm{\mathcal{U}}_{F(s)}},
\end{aligned}
\label{eqn:rec_derivative}
\end{equation}
where  $\mu_s=\min(s), \nu_s=\max(s)$. Notice that when $F(s)$ is equal to node $D$, the above recursive procedure terminates.

Furthermore, with Eq. (\ref{eqn:rec_derivative}) the gradients for leaf frames and transfer tensors can also be recursively calculated as: 
\begin{align}
\frac{\partial L}{\partial\bm{\mathcal{U}}_s}=&\frac{\partial\bm{\mathcal{Y}}}{\partial\bm{\mathcal{U}}_s}\times_{1,\cdots,\mu_s-1,\nu_s+1,\cdots,d}^{\nu_s-\mu_s+3,\cdots,d+1}\frac{\partial L}{\partial\bm{\mathcal{Y}}},\label{eqn:gradients_u}\\
\frac{\partial L}{\partial\bm{\mathcal{B}}_s}=&\frac{\bm{\mathcal{Y}}}{\partial\bm{\mathcal{U}}_s}\times_{2,\cdots,\nu_{s_1}-\mu_{s_1}+2}^{2,\cdots,\nu_{s_1}-\mu_{s_1}+2} \bm{\mathcal{U}}_{s_1}\nonumber\\
&\times_{2,\cdots,\nu_{s_2}-\mu_{s_2}+2}^{3,\cdots,\nu_{s_2}-\mu_{s_2}+3}\bm{\mathcal{U}}_{s_2}\times_{1,\cdots,d}^{4,\cdots,d+3}\frac{\partial L}{\partial \bm{\mathcal{Y}}}.\label{eqn:gradients_g}
\end{align}

In general, our proposed ``integrate-then-decompose" compression strategy can bring significant performance benefit for the FDHT-LSTM models. As shown in Figure \ref{fig:full_compress}, compared with the naive layer-wise compression strategy that suffers significant accuracy degradation, our proposed approach ensures high accuracy without performance loss. Furthermore, compared to the input-to-hidden-only compression counterpart, our FDHT-LSTM enjoys much smaller model size. More detailed and comprehensive empirical evaluation results across different datasets will be reported in Section \ref{sec:eval}.


\section{The Proposed FDHT-LSTM: Hardware Architecture}
\label{sec:hw}

Based on the above described FDHT-LSTM structure and algorithm, in this section we further develop the corresponding hardware architecture that can fully leverage the algorithmic benefits to support the efficient execution of the FDHT-LSTM model.

\subsection{Analysis of Computation Flow: A 2-D Matrix Perspective}
\begin{figure*}[t]
\centering
\includegraphics[width=6in]{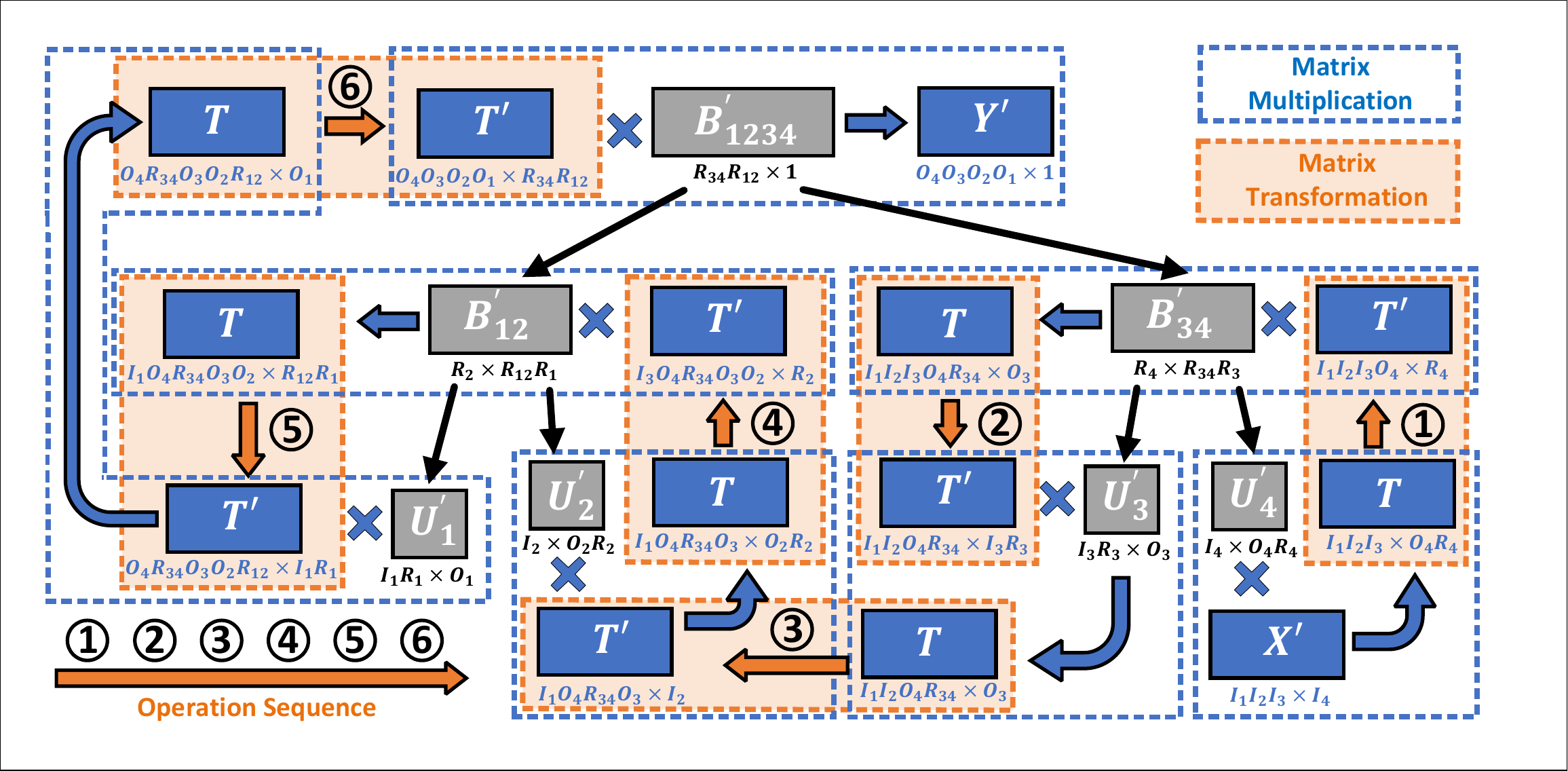}%
\caption{Computation flow of high-order FDHT-LSTM from the perspective of 2-D matrix multiplication and transformation.}
\label{fig:Computation-flow}
\end{figure*}

As described in Section \ref{sec:alg}, the overall computation scheme of FDHT-LSTM can be interpreted as a series of tensor contraction along a binary tree (see Figure \ref{fig:htl}). Consider tensor contraction (Eq. \ref{eqn:contraction}) is essentially a high-dimensional operation; while the underlying arithmetic and memory hardware components can only support 2-D operation, a re-analysis of the computational flow, from 2-D matrix perspective, is very necessary and desired. 

Figure \ref{fig:Computation-flow} illustrates the overall computational flow of the proposed FDHT-LSTM in the 2-D matrix format. Here $\bm{B}^{'}_{1234}$, $\bm{B}^{'}_{12}$, $\bm{B}^{'}_{34}$, $\bm{U}^{'}_{1}$, $\bm{U}^{'}_{2}$, $\bm{U}^{'}_{3}$ and $\bm{U}^{'}_{4}$ are the weight matrices of the flatten high-order tensors, $\bm{X^{'}}$ is the reshaped input matrix data, and 
$\bm{T}$ is the intermediate result. Notice that here $\bm{T}$ cannot be multiplied with the next weight matrix directly due to the mismatch of matrix dimension. Therefore, $\bm{T}$ needs to be first transformed to $\bm{T}^{'}$ to ensure functional validity.

As shown in Figure~\ref{fig:Computation-flow}, a total of six matrix transformations (from $\bm{T}$ to $\bm{T^{'}}$) is required in the entire computation flow of FDHT-LSTM. After further analysis, we can obtain two important observations: \ul{First}, these transformations are not simple matrix reshape but essentially  permutation that requires complex matrix operations such as row/column merge and concatenation etc.; \ul{Second} the six transformations can be categorized to three types. To be specific, \textbf{Type-I} (Transformation \ballnumber{1} and \ballnumber{4}) is to convert $\bm{T}\in\mathbb{R}^{A\times (B_1\times B_2)}$ to $\bm{T^{'}}\in\mathbb{R}^{(A\times B_1)\times B_2}$, and the mapping principle for this transformation is as:
\begin{align}
\label{eqn:resh.14}
    \begin{split}
        \bm{T}&(m,n) \Rightarrow \bm{T^{'}}(p, q),
    \end{split}
\end{align}
where $m = a$, $n = (b_1-1)\times B_2+b_2$, $p = (a-1) \times B_1 +  b_1$, $q = b_2$, $a = 1, 2,..., A$, $b_1 = 1, 2,..., B_1$ and $b_2 = 1, 2,..., B_2$.

\textbf{Type-II} (Transformation \ballnumber{2} and \ballnumber{5}) is to convert $\bm{T}\in\mathbb{R}^{(A_1\times A_2 \times A_3) \times (B_1\times B_2)}$ to $\bm{T^{'}}\in\mathbb{R}^{(A_1\times A_3\times B_1)\times (A_2\times B_2)}$, and the mapping principle for this transformation is as:
\begin{align}
\label{eqn:resh.25}
    \begin{split}
        \bm{T}&(m, n) \Rightarrow  \bm{T^{'}}(p, q),
    \end{split}
\end{align}
where $m = (a_1-1)\times A_2 \times A_3 + (a_2 - 1) \times A_3 + a_3$, $n = (b_1-1)\times B_2+b_2$, $p = (a_1-1)\times A_3 \times B_1 + (a_3 - 1) \times B_1 + b_1$, $q = (a_2 -1) \times b_2$, $a_1=1, 2,..., A_1$, $a_2=1, 2,..., A_2$, $a_3=1, 2,..., A_3$, $b_1=1, 2,..., B_1$ and $b_2=1, 2,..., B_2$.

\textbf{Type-III} (Transformation \ballnumber{3} and \ballnumber{6}) is to convert $\bm{T}\in\mathbb{R}^{(A_1\times A_2\times A_3)\times B}$ to $\bm{T^{'}}\in\mathbb{R}^{(A_1\times A_3\times B)\times A_2}$, and the mapping principle for this transformation is as:
\begin{align}
\label{eqn:resh.36}
    \begin{split}
        \bm{T}&(m, n) \Rightarrow \bm{T^{'}}(p, q),
    \end{split}
\end{align}
where $m = (a_1-1)\times A_2 \times A_3 + (a_2 - 1) \times A_3 + a_3$, $n =  b$, $p = (a-1) \times A_3 \times B + (a_3-1)\times B + b$, $q = a_2$, $a_1=1, 2,..., A_1$, $a_2=1, 2,..., A_2$, $a_3=1, 2,..., A_3$ and $b=1, 2,..., B$.

\begin{figure*}
    \centering
    \includegraphics[width=6.0in]{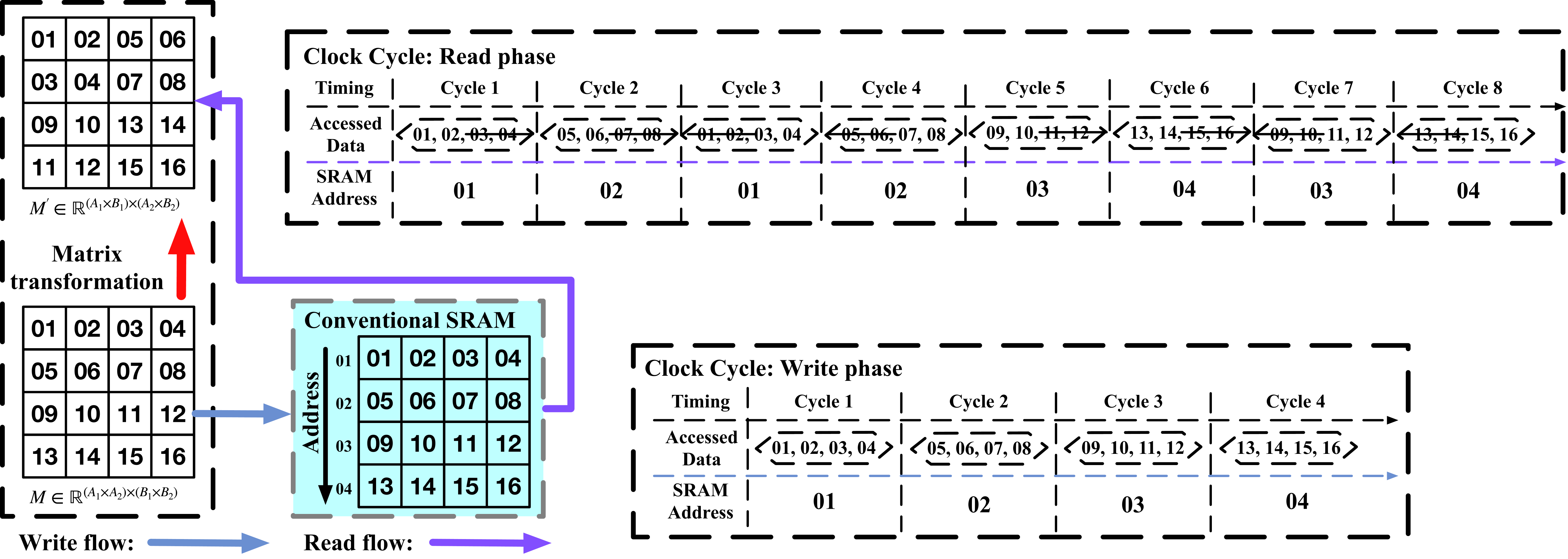}
    \caption{Data access in conventional SRAM architecture.}
    \label{fig:data_access}
\end{figure*}

\subsection{Access Conflict on Conventional SRAM Architecture}
Without losing the generality, we use the basic matrix transformation  $\bm{M}\in\mathbb{R}^{(A_1\times A_2) \times (B_1\times B_2)} \Rightarrow  \bm{M}^{'}\in\mathbb{R}^{(A_1\times B_1) \times (A_2\times B_2)}$ to elaborate the challenges incurred by the complicated transformation above, where $A_1=A_2=B_1=B_2=2$. As shown in Figure \ref{fig:data_access}, when using conventional SRAM architecture, in the writing phase, each row of the matrix is written to one address (one row) in this SRAM in one cycle. For example, data $01, 02, 03$ and $04$ is written to the address $01$ of the SRAM. \textbf{However, the access conflict happens in the read phase}. When reading the SRAM, the desired data is $01, 02$ and $04, 05$. These four data are located at two different addresses $01$ and $02$, and these two addresses can not be accessed simultaneously, thereby causing access conflict. 

The basic solution to the access conflict problem is relocating one address multiple times (as shown in Figure \ref{fig:data_access}). More specifically, in cycle 1, address $01$ is located and data $01, 02, 03, 04$ is read out from the SRAM. Data $01, 02$ is saved in the register file, while $03, 04$ is the undesired data and discarded. Similarly, in cycle 2, address $02$ is located; $05, 06, 07, 08$ is read out from SRAM, and $05, 06$ is stored in the register file and $07, 08$ is discarded. In the assemble unit, $01, 02$ and $05, 06$ form the data array $01, 02, 05, 06$. \textbf{However, such a solution suffers two main limitations:}

1. Additional memory overhead is needed to support matrix transformation. Since the matrix transformation cannot be executed in an on-the-fly way when using conventional memory architecture, an additional register file is necessary to save the desired data. Thus, this solution significantly increases the cost of the register file, reducing the area and energy performance of the overall hardware.

2. The write and reading scheme becomes much more complicated because of frequent re-locating of the memory address. Also, notice that the example we raise here is only a simple case, and in practical workload, the matrix transformation is very large and complicated. Hence it is not feasible to read all the data to register files and assemble them when using conventional SRAM architecture.

\subsection{2-D SRAM Array}

\begin{figure}[t]
    \centering
    \includegraphics[width=\columnwidth]{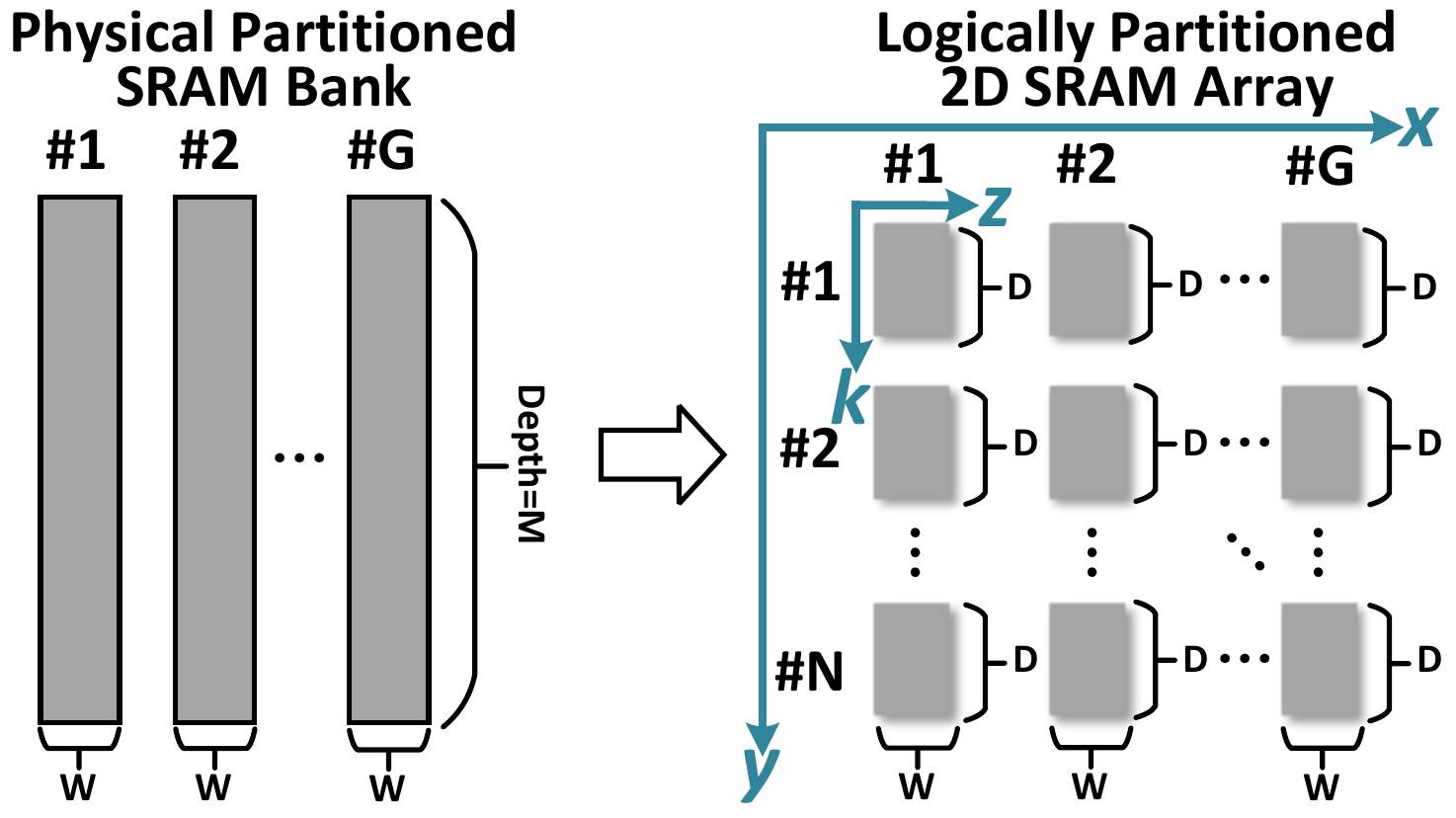}
    \caption{The physical SRAM bank and logically partitioned 2-D SRAM array. Here $W$ and $M$ are the width and depth of one SRAM bank, respectively. $D$ is the depth of one segment of SRAM bank, where $M=D\times N$. Here $D$ can be reconfigured according to workload.}
    \label{fig:2-D-SRAM-array}
\end{figure}

In order to implement the desired matrix transformation, which is essentially the 2-D mapping of high-order \textit{tensor transpose}, we propose to read and write the data in the \textit{2-D SRAM array}. In other words, we aim to increase the dimension of the SRAM from physical 2-D to "virtually" 4-D, to better support the operation in the high-order tensor operation. As illustrated in Figure \ref{fig:2-D-SRAM-array}, our proposed SRAM array \textit{physically} consists of $G$ banks of SRAM, where each of them has width of $W$ and depth of $M$. Based on such fixed arrangement, the controller will \textit{logically} partition the SRAM array to more fine-grained format when supporting different workloads with various matrix sizes. To be specific, according to the demand of the workload, each SRAM bank is logically partitioned into $N$ \textit{segments} along the depth dimension, and each segment has a depth of $D$ where $M=N\times D$. Notice that here $G$, $M$ and $W$ are the fixed dimensions of the entire SRAM module, and $D$ and $N$ are reconfigurable and determined by the main controller, thereby providing the controller sufficient capability to adjust the granularity of the memory operation for different workloads. 

To achieve this goal without introducing any memory access conflict, delicate design of memory read/write scheme is required and will be described next.

\subsection{Matrix Transformation on 2-D SRAM Array}
\label{subsec:basic trans}

Based on the 2-D SRAM array, the corresponding conflict-free memory access scheme can be developed. Without loss of generality, in this subsection we first study the read and write schemes for a basic transformation  $\bm{M}\in\mathbb{R}^{(A_1\times A_2) \times (B_1\times B_2)} \Rightarrow  \bm{M}^{'}\in\mathbb{R}^{(A_1\times B_1) \times (A_2\times B_2)}$. The obtained outcome will further serve as the foundation of implementing the specific  matrix transformations (Eq.\ref{eqn:resh.14} -- Eq.\ref{eqn:resh.36}) used in FDHT-LSTM.   

To facilitate the notation, as illustrated in Figure \ref{fig:2-D-SRAM-array}, the bank and segment indices of the 2-D SRAM array are denoted as \textit{x} and \textit{y}, respectively, where $x=1, 2,..., G$ and $y=1, 2,..., N$. In addition, \textit{z} and \textit{k} are the row and depth indices in each SRAM segment, respectively, where $z=1, 2, ..., W$ and $k=1, 2,..., D$. Following such notation, one data entry in one SRAM segment, one row in one SRAM segment, one SRAM segment, and one SRAM bank can be represented as $S(x, y, k, z)$, $S(x, y, k, :)$, $S(x, y, :, :)$ and $S(x, :, :, :)$, respectively. Also, we use $\bm{M}(a, :)$ and $\bm{M}(:, b)$ to represent the $a$-th row vector and the $b$-th column vector of $\bm{M}$, respectively. In addition, $\bm{M}(a,b:c)$ and $\bm{M}(a:b,c)$ denote the part of $\bm{M}(c,:)$ and $\bm{M}(:, c)$ that are with column and row indices from $a$ to $b$.






\begin{figure*}
    \centering
    \includegraphics[width=7.0in]{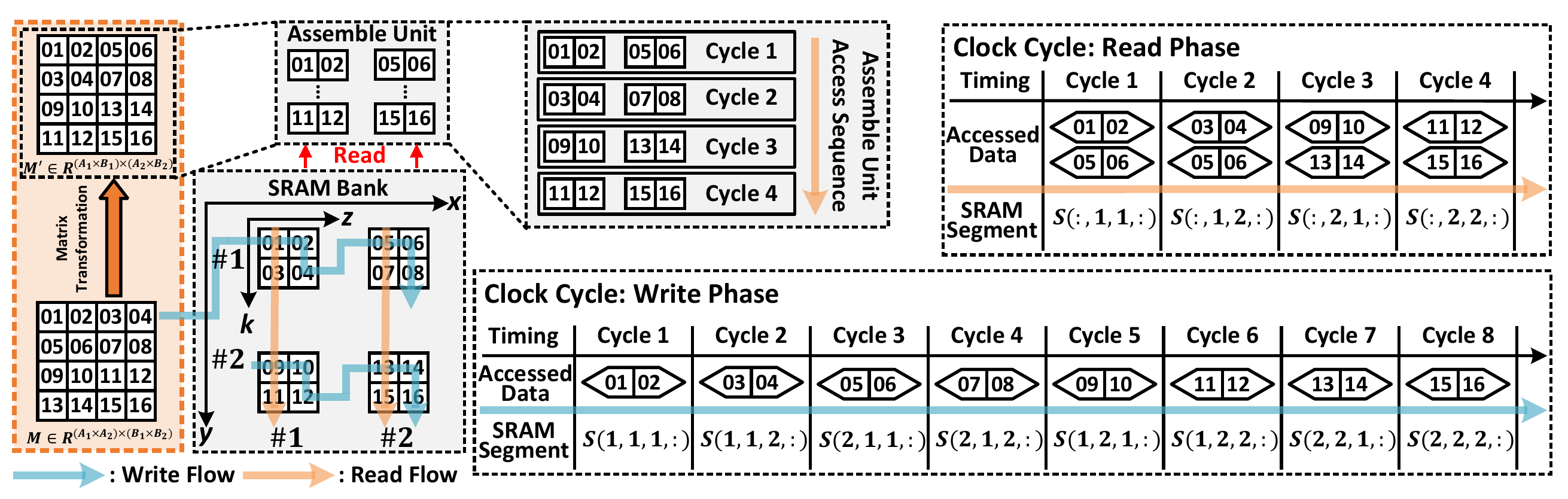}
    \caption{The proposed write and read scheme to 2-D SRAM array for a basic matrix transformation.}
    \label{fig:dataflow}
\end{figure*}

\subsubsection{Memory Write Scheme}
Next we describe memory write scheme for the example transformation $\bm{M}\in\mathbb{R}^{(A_1\times A_2) \times (B_1\times B_2)} \Rightarrow  \bm{M}^{'}\in\mathbb{R}^{(A_1\times B_1) \times (A_2\times B_2)}$. Here $D$, as the granularity of 2-D SRAM array, is configured as $B_2$ to support the operation for this example. Initially, $\bm{M}(1,1:B_2)$ is written along the $z$ dimension into $S(1, 1, 1,:)$ as the first row of SRAM segment $S(1, 1)$.
Then, if we interpret the $B_2$ data entries in one row as a group, and one row of $\bm{M}$ contains $B_1$ groups, these $B_1$ data groups are scheduled to be sequentially written into $S(1, 1, :, :)$ in a row-wise way. For instance, the data entries of $\bm{M}(1,B_2+1: 2\times B_2)$ are written into $S(1, 1, 2,:)$, and the data entries of $\bm{M}(1,(B_1-1)\times B_2+1 : B_1\times B_2)$ are written into $S(1, 1, B_1, :)$.
After that, a row of data entries of $\bm{M}$ is stored in one segment of 2-D SRAM array.
Then, $A_2$ rows of $\bm{M}$ are written into $A_2$ SRAM segments along the \textit{x} dimension, e.g., $S(1, 2, :, :)$ stores the data of $\bm{M}(2, :)$ and $S(1, S_2, :, :)$ stores the data of $\bm{M}(A_2, :)$.
By using this way, $\bm{M}(1:A_2,:)$ is written into one SRAM bank $S(1,:,:,:)$. Writing the rest of the data in $\bm{M}$ to SRAM follows the similar way, e.g., $S(2,:,:,:)$ stores the data of $\bm{M}(A_2+1:2\times A_2,:)$, and $S(A_1,:,:,:)$ stores the data of $\bm{M}((A_1-1)\times A_2+1: A_1\times A_2,:)$. Figure \ref{fig:dataflow} shows the detailed scheme for an example matrix transformation, where $G=2$, $M=4$, $W=2$, and $A_1=A_2=B_1=B_2=2$. It is seen that with such writing scheme the 2-D SRAM array stores the re-arranged data, which will be ready to realize the desired transformation after proper reading scheme.

\subsubsection{Memory Read Scheme}
In the reading phase, the data in $G$ banks are read simultaneously in a row-wise way. To be specific, the data in $S(:, 1, :, k)$ are first read in the first cycle, and then they are sent to assemble unit to form $\bm{M^{'}}(1, :)$. In the following cycles, such operation is repeated row by row along the \textit{k} and \textit{y} dimensions. Figure \ref{fig:dataflow} shows the details of reading scheme for the example matrix transformation.

In general, the proposed 2-D SRAM with a corresponding write and read scheme can solve the challenges mentioned above:

1. The matrix transformation can be executed in an on-the-fly way. In the reading phase, all the to-be-read data is the desired data and only a small register file is needed to save the data. In our FDHT-LSTM, the size of the register file is only $448 Bytes$.  

2. Since in the writing and reading phase, the desired memory address increases sequentially, it is very easy to design and implement the address controller. Also, because the 2-D SRAM architecture is logically split, it is very easy to be reconfigured and very flexible for different matrix transformations with various shapes.

\subsection{Realizing Matrix Transformation for FDHT-LSTM}
\label{subsec:spec_trans}

Recall that the read/write access schemes described in Section \ref{subsec:basic trans} are designed for a basic matrix transformation that is different from the ones described in Eq. \ref{eqn:resh.14} -- Eq. \ref{eqn:resh.36}. In this subsection we leverage the approach developed in Section \ref{subsec:basic trans} to further realize the three matrix transformations specifically adopted in the computation flow of FDHT-LSTM.


\subsubsection{Memory Access Scheme for Type-I Transformation}
\begin{figure}
    \centering
    \includegraphics[width=\columnwidth]{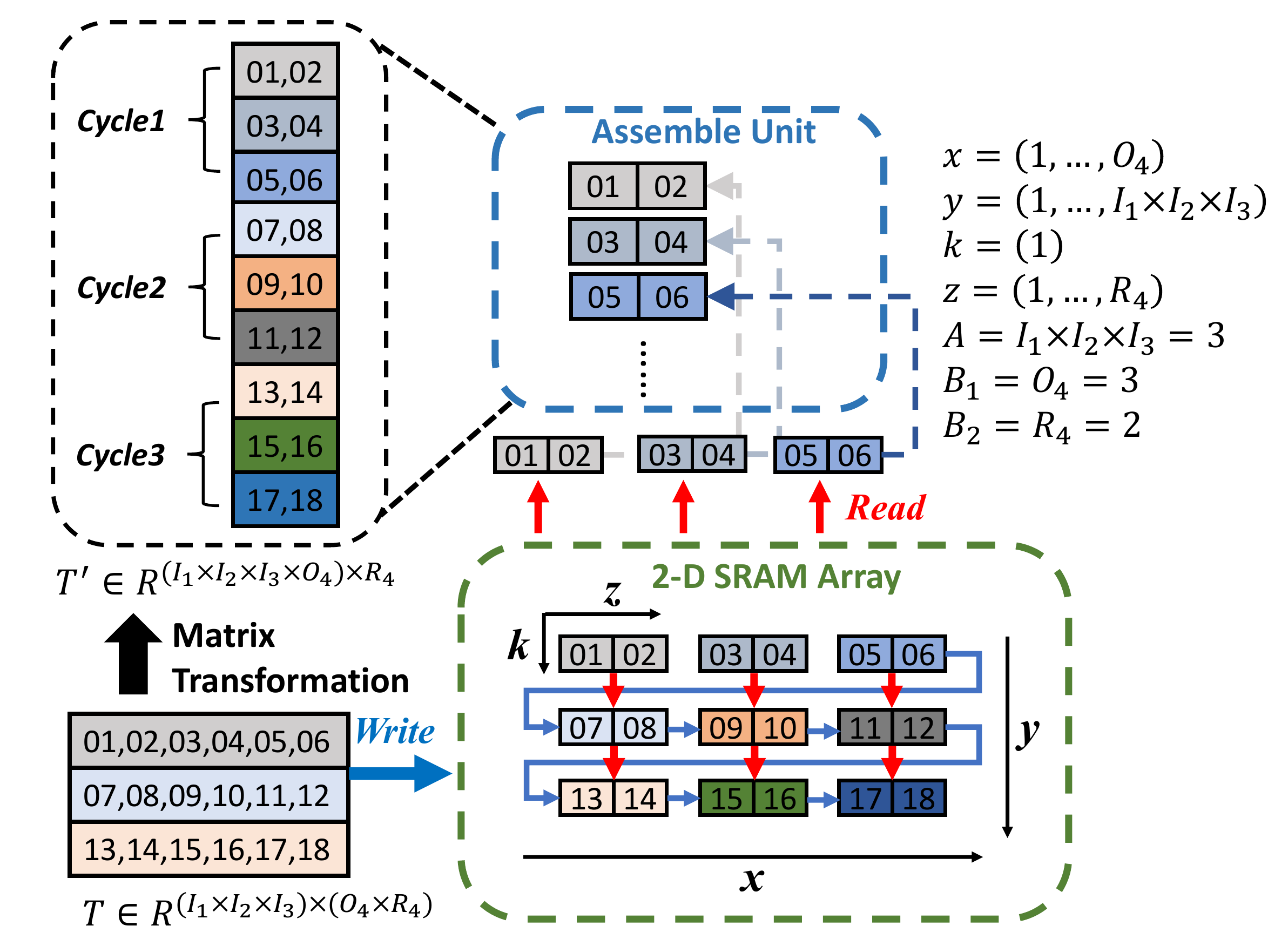}
    \caption{Example memory write and read schemes for Transformation 1.}
    \label{fig:mat-reshaping-1}
\end{figure}

In this case $D$ is set as 1. During the memory writing phase, consider $B_2$ data as a group, then the data in each group are written to one row of SRAM segment along the \textit{z} dimension. Since each row of $\bm{T}$ contains $B_1$ data groups, they are written into one row of SRAM array along the \textit{x} dimension sequentially.
In other words, one row of SRAM array $S(:,1,:,:)$ stores one row of $\bm{T}$ as $\bm{T}(1, :)$.
Following this strategy, the $A$ rows of $\bm{T}$ are written into the SRAM array row by row along the \textit{y} dimension in a sequential way. Then during the memory reading phase, consider each row of SRAM array $S(:,1,:,:)$ has $B_1 \times B_2$ data; hence the assemble unit will partition one row data to form a data array with $B_1$ columns and $B_2$ rows. By repeating such reading operation along the $y$ dimension, $\bm{T^{'}}$ can be read out and sent to PE array for further processing. Figure~\ref{fig:mat-reshaping-1} illustrates the details of memory write and read schemes to an example Transformation \ballnumber{1}, where $A=I_1\times I_2 \times I_3=3$, $B_1=O_4=3$ and $B_2=R_4=2$.

\subsubsection{Memory Access Scheme for Type-II Transformation} 
\begin{figure}
    \centering
    \includegraphics[width=\columnwidth]{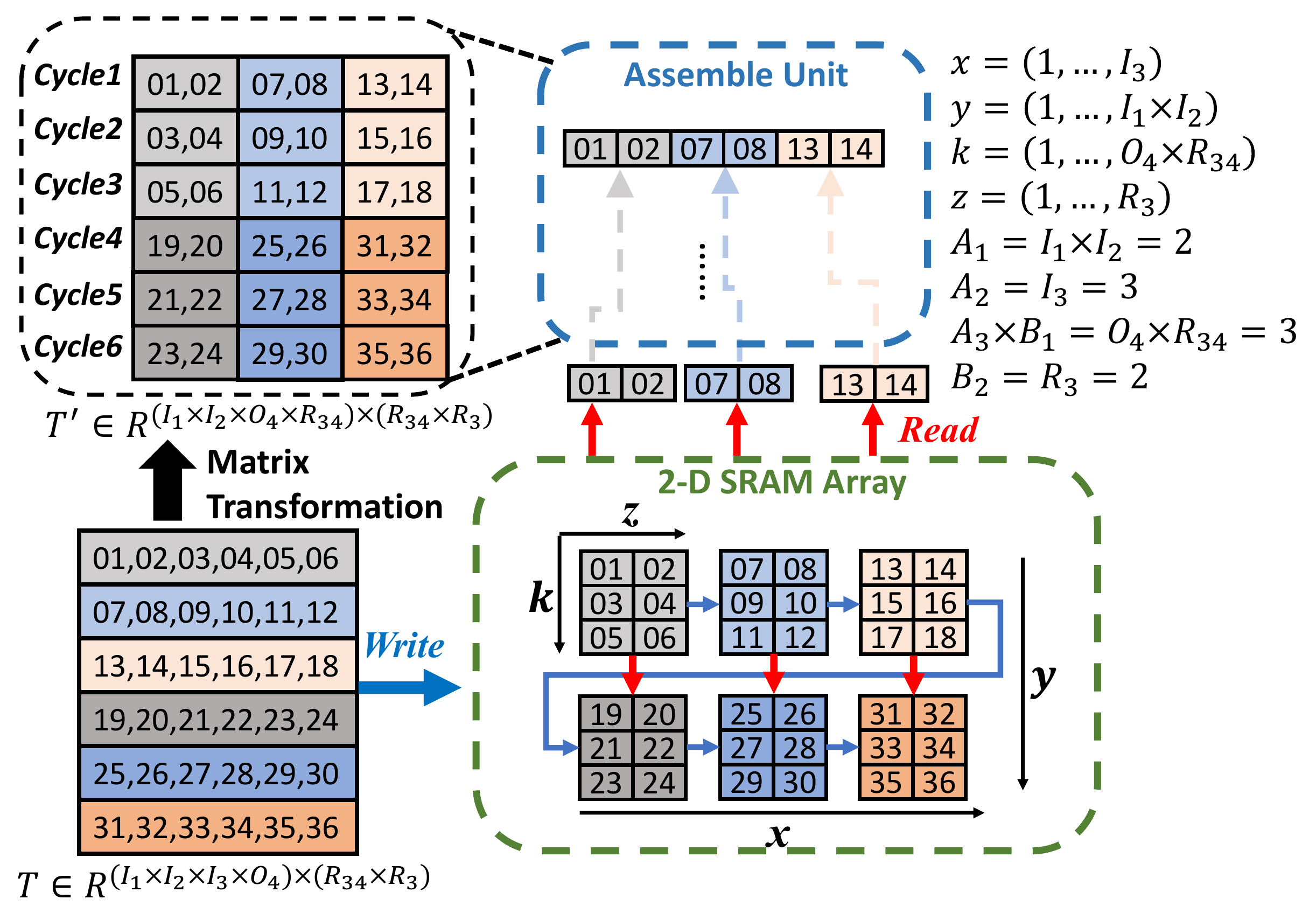}
    \caption{Example memory write and read schemes for Transformation 2.}
    \label{fig:mat-reshaping-2}
\end{figure}
In this case, the transformation can be rewritten as $\bm{T}\in\mathbb{R}^{(A_1\times A_2) \times ((A_3\times B_1) \times B_2)} \Rightarrow \bm{T^{'}}\in\mathbb{R}^{(A_1\times (A_3\times B_1))\times (A_2\times B_2)}$, and then it becomes the basic format example described in Section \ref{subsec:basic trans} with $D=A_3\times B_1$.
Then, during the writing phase, we rearrange the data of $\bm{T}$ along $B_2$, $A_3\times B_1$, $A_2$ and $A_1$ dimensions on the \textit{z}, \textit{k}, \textit{x} and \textit{y} dimensions of 2-D SRAM array, respectively. In the memory reading phase, $A_2 \times B_2$ data entries in the SRAM array are read simultaneously and combined as one row of $\bm{T^{'}}$. In general, by reading along the $k$ and $y$ dimensions of 2-D SRAM array, the row data of $\bm{T^{'}}$ as $A_1\times A_3\times B_1$ entries can be obtained. Figure~\ref{fig:mat-reshaping-2} illustrates the details of memory access scheme to an example Transformation \ballnumber{2}, where $B_2=R_3=2$, $A_3 \times B_1 = O_4\times R_{34}=3$, $A_2 = I_3 = 3$ and $A_1 = I_1\times I_2 = 2$.

\subsubsection{Memory Access Scheme for Type-III Transformation}
\begin{figure}
    \centering
    \includegraphics[width=\columnwidth]{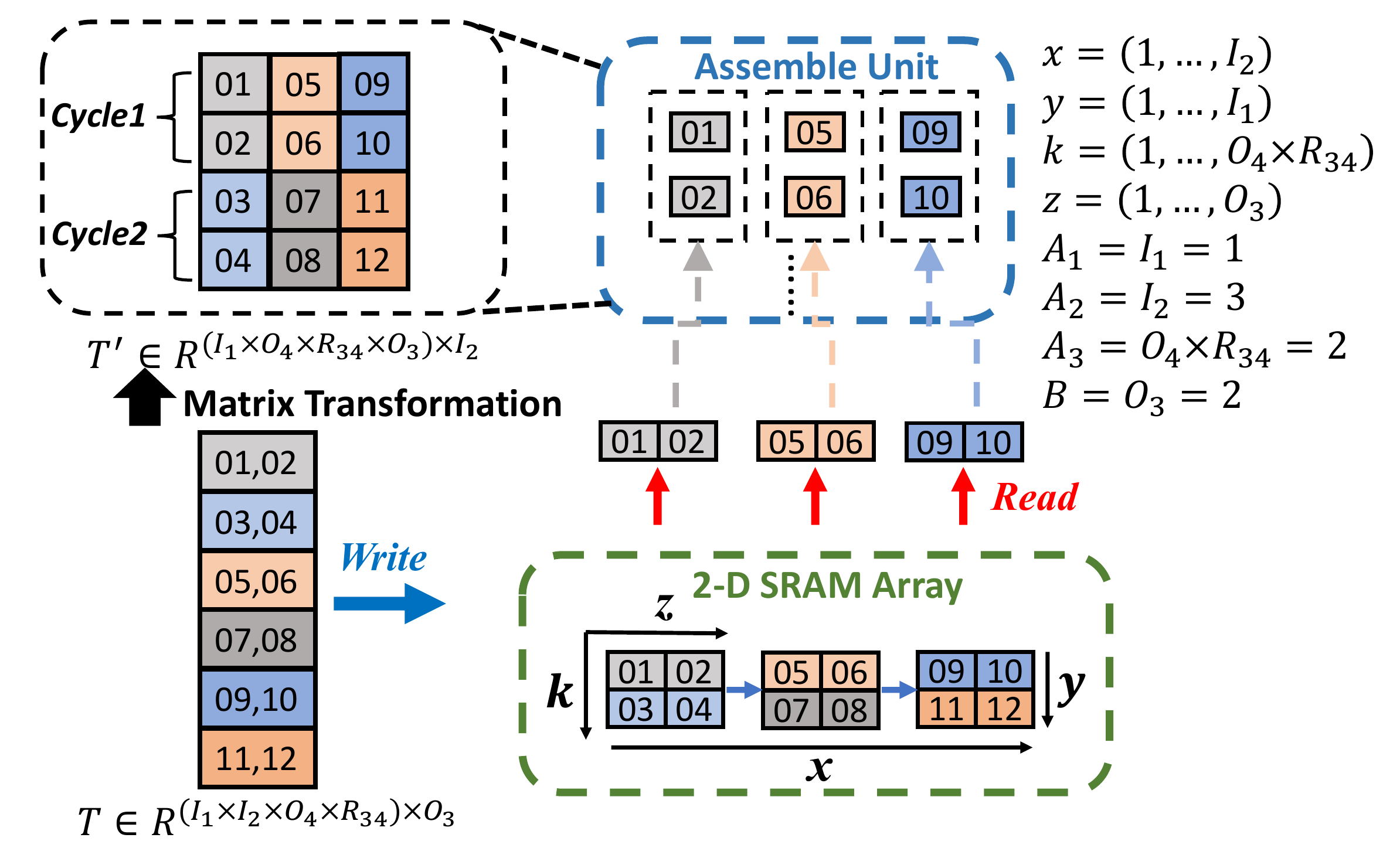}
    \caption{Example memory write and read schemes for Transformation 3.}
    \label{fig:mat-reshaping-3}
\end{figure}

In the writing phase of this case, consider each row of $\bm{T}$ as a group, and the data in each group is written along the dimension \textit{z} in each SRAM segment, e.g., $S(1,1,1,:)$ stores the data in $\bm{T}(1,:)$. 
Consequently, in order to keep $A_3$ in the column index of $\bm{T^{'}}$, $D$ should be set as $A_3$ to ensure that each $A_3$ rows of $\bm{T}$ are written along the $k$ dimension in each SRAM segment, e.g., $\bm{T}(1:A_3,:)$ is stored in $S(1, 1, :, :)$.
In addition, the data of $\bm{T}$ along $A_2$ and $A_1$ dimensions are re-arranged on the \textit{x} and \textit{y} dimensions of 2-D SRAM array, respectively, to guarantee $A_2$ appear in the row index of $\bm{T}^{'}$.
During the reading phase, $A_2 \times B$ data entries in the SRAM array are read simultaneously, and 
they are reshaped into an array with $A_2$ columns and $B$ rows in the assemble units. Figure~\ref{fig:mat-reshaping-3} illustrates the details of memory read and write schemes of the Transformation \ballnumber{3}, where $B=O_3=2$, $A_3=O_4\times R_{34}=2$, and $A_1=I_1=1$.

\subsection{Overall Architecture}
\begin{figure}[t]
    \centering
    \includegraphics[width=\columnwidth]{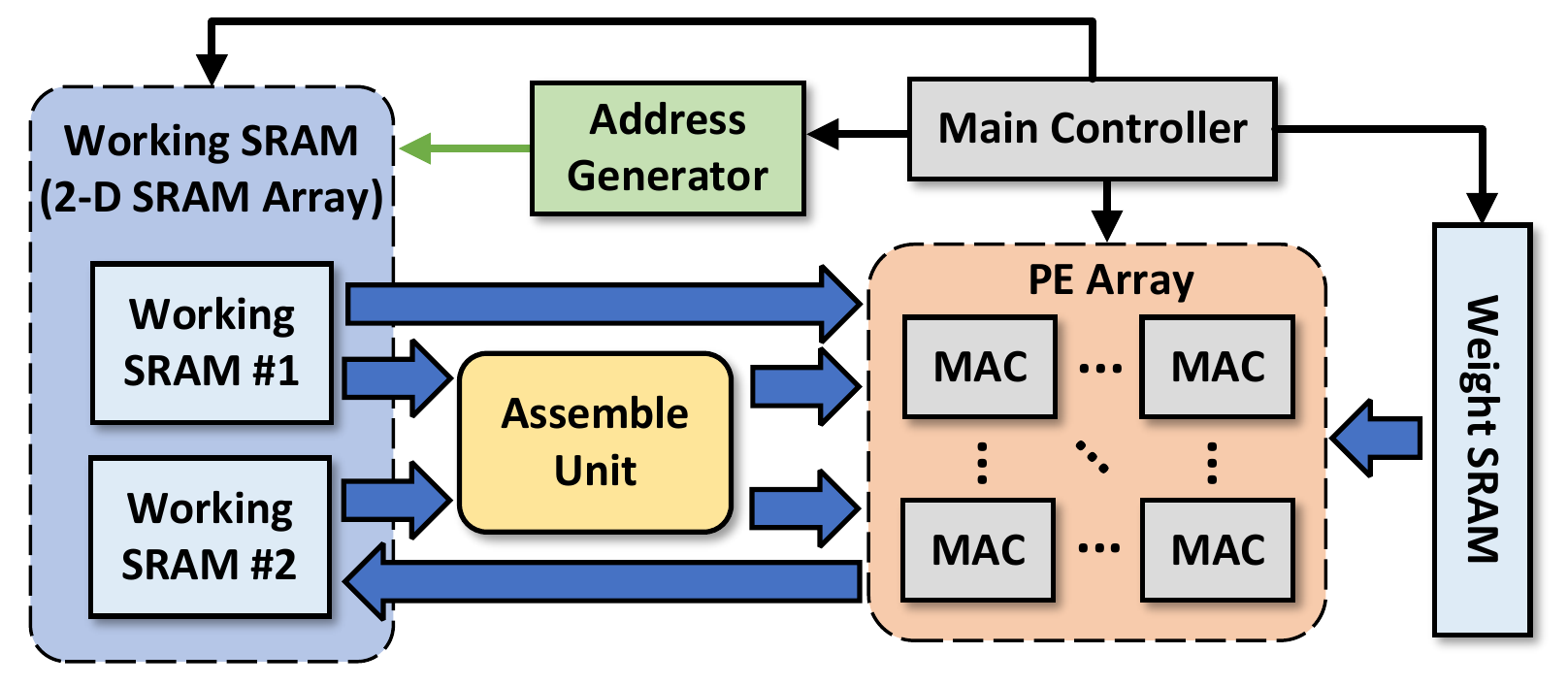}
    \caption{The overall hardware architecture.}
    \label{fig:Overall-architecture}
\end{figure}

Based on the proposed 2-D SRAM array and the corresponding read/write access schemes, the hardware architecture that supports the execution of FDHT-LSTM models can be developed. As illustrated in  Figure~\ref{fig:Overall-architecture}, the datapath of the accelerator is a PE Array that consists of multiply-accumulators (MACs) to perform matrix multiplication. \textcolor{black}{Notice that because FDHT-LSTM operates on the the dense matrices rather than the sparse ones, the multiplier utilization is very high.} In addition, the weight of the factorized tensor cores and the intermediate results are stored in the weight SRAM and working SRAM, respectively. Notice that here only working SRAM is in the format of 2-D array since the desired matrix transformation is only involved with the intermediate results. The read/write access schemes described in Section \ref{subsec:spec_trans} are mapped into an individual address generator, which receives the configuration information from main controller for different workloads.

\section{Evaluation}
\label{sec:eval}

\subsection{Performance of FDHT-LSTM Model}
We evaluate the performance of our proposed FDHT-LSTM approach on various video recognition tasks and compare it with other state-of-the-art tensor decomposition-based LSTM compression methods. In particular, in order to verify the benefits of our full decomposition strategy and fairly compare the HT decomposition with other tensor decomposition for LSTM compression, we also conduct experiments on only performing HT decomposition on the input-to-hidden layer of LSTM models.

\subsubsection{Experimental Settings}

For fair comparison, we set $d$, the order of the target tensors of the tensorized input, output and weights, identical to the settings in prior works \cite{yang2017tensor} and \cite{ye2018learning}. For the tensor ranks, we also follow the prior works to select $r$ towards reaching a good balance between compression ratio and model accuracy. For the setting of other hyper-parameters, dropout rate and batch size are set as 0.25 and 16, respectively, and ADAM is selected as the underlying training optimizer with weight decay of 0.001.



\begin{table}
\normalsize
\caption{Compression settings for HT-layer. Here because YTC dataset is smaller than UCF11 dataset, we choose smaller $R$ (non-leaf) for LSTM-YTC to achieve higher compression ratio with still preserving high accuracy.}
\def\arraystretch{1.05}
\centering
\begin{tabular}{r||c|c} 
\hline

\textbf{Model}  &      LSTM-UCF11 &LSTM-YTC \\
\hline
\textbf{Size} & $(57600, 256)$ & $(57600, 256)$\\
$d$ & $4$ & $4$\\
$I$ & $[16, 16, 16, 15]$ & $[16, 16, 16, 15]$\\
$O$ & $[4, 4, 4, 4]$ & $[4, 4, 4, 4]$\\
$R$ \textbf{(Leaf)} & $14$ & $14$ \\
$R$ \textbf{(Non-leaf)} & $12$ & $11$\\\hline
\textbf{Compr. Ratio} & 6,726$\times$&7,117$\times$\\
\hline
\end{tabular}
\label{tab:exp-settings}
\end{table}



\subsubsection{Results on UCF11 Dataset} 

The UCF11 dataset \cite{liu2009recognizing} is a set of 1,600 human action video clips which can be categorized as 11 classes. The resolution of each clip is $320\times 240$. We follow the prior works \cite{ye2018learning} \cite{pan2019compressing} to pre-process the video data. To be specific, the resolution of each video clip is downgraded to $160 \times 120$, then we randomly sample 6 frames from each clip and combine them as a single sequential data point. In the original uncompressed LSTM model, there are 4 input-to-hidden layers and 4 hidden-to-hidden layers. We also follow the prior works to set the input size and the number of hidden states in the baseline model as $160\times 120\times 3=57,600$ and 256, respectively. For the compressed FDHT-LSTM model, the concatenated vector $\bm{I}$ can be reshaped as a tensor of shape as $16\times 16\times 16\times 15$, and the hidden state is reshaped to a tensor of shape as $4\times 4\times 4\times 4$. In addition, all leaf and non-leaf ranks are set as 14 and 12, respectively.


\begin{table}[t]
\normalsize
\caption{Compression performance of different tensor decomposed-based LSTM models on UCF11 dataset.}
\centering
\def\arraystretch{1.05}
\begin{tabular}{c|c|c|c}
\hline
\multirow{2}{*}{\textbf{Model}}  & \multicolumn{2}{c|}{\textbf{Number of Parameters}}   & \textbf{Top-1} \\
\cline{2-3}
 & \textbf{Input-to-Hidden} & \textbf{Overall} & \textbf{Acc.} (\%) \\
\hline
\hline
LSTM  & 58.98M     & 59.24M  & 69.7     \\
\hline
TT-LSTM \cite{yang2017tensor}  & \multirow{2}{*}{3,360} & 265.50K & \multirow{2}{*}{79.6}    \\ (ICML'17)
& & (223$\times$) &\\
\hline
BT-LSTM \cite{ye2018learning}  & \multirow{2}{*}{3,387}  & 265.53K  & \multirow{2}{*}{85.3}     \\
(CVPR'18) & & (223$\times$) &\\
\hline
TR-LSTM \cite{pan2019compressing}  & \multirow{2}{*}{1,725}  & 263.87K & \multirow{2}{*}{86.9}     \\
(AAAI'19) & & (225$\times$) & \\
\hline
HT-LSTM  & \multirow{2}{*}{1,245}  & 263.39K & \multirow{2}{*}{87.2}     \\
(Ours) & & (225$\times$) & \\
\hline
\hspace{-1mm}\textbf{FDHT-LSTM}\hspace{-1mm}   & \multirow{2}{*}{N/A} & \textbf{8,808} & \multirow{2}{*}{\textbf{87.5}}    \\
\textbf{(Ours)} & & (\textbf{6,726$\times$}) & \\
\hline
\end{tabular}

\label{tab:direct_ucf11}
\end{table}

The compression performance of the proposed FDHT-LSTM and other state-of-the-art tensor decomposed LSTM are summarized in Table \ref{tab:direct_ucf11}. It is seen that compared with the original uncompressed LSTM model with 59 million parameters, our proposed FDHT-LSTM only needs 8,808 parameters while resulting in a 17.8\% accuracy increase. Compared with other tensor decomposition-based LSTM compression methods, i.e., TT-LSTM, BT-LSTM and TR-LSTM, FDHT-LSTM enjoys order-of-magnitude reduction in model parameters and at least 0.6\% higher accuracy.

\subsubsection{Results on Youtube Celebrities Face Dataset} 

Youtube celebrities face dataset \cite{kim2008face} contains 1,910 video clips with different resolutions which are categorized as 47 labels. Similar to the case of UCF11 dataset, we also follow the prior works \cite{ye2018learning} \cite{pan2019compressing} for data pre-processing. To be specific, we scale down all the video clips to $160\times 120$ and randomly sample 6 frames to form a single sequence. Other experimental settings are also consistent with the case of UCF11 dataset except that all the non-leaf ranks are set as 11.

Table \ref{tab:directly_ytc} summarizes the compression performance on this dataset for different tensor decomposed LSTMs. It is seen that compared to the baseline LSTM model our FDHT-LSTM can provide as high as 7,117$\times$ compression ratio with test accuracy increase. Compared with the state-of-the-art TT-LSTM, our FDHT-LSTM still enjoys 6,894$\times$ smaller model size and 12.7\% higher accuracy. 

\begin{table}[t]
\normalsize
\caption{Compression performance of different tensor decomposed-based LSTM models on Youtube celebrities face dataset.}
\def\arraystretch{1.05}
\centering
\begin{tabular}{c|c|c|c}  
\hline
\multirow{2}{*}{\textbf{Model}}  & \multicolumn{2}{c|}{\textbf{Number of Parameters}}   & \textbf{Top-1} \\
\cline{2-3}
 & \textbf{Input-to-Hidden} & \textbf{Overall} & \textbf{Acc.} (\%) \\
\hline
\hline
LSTM &  58.98M  & 59.24M  & 33.2       \\
\hline
TT-LSTM \cite{yang2017tensor} & \multirow{2}{*}{3,392}  & 265.54K  & \multirow{2}{*}{75.5}      \\
(ICML'17) & & (223$\times$) & \\
\hline
HT-LSTM & \multirow{2}{*}{810}  & 262.95K  & \multirow{2}{*}{88.1}      \\
(Ours) & & (225$\times$) & \\
\hline
\hspace{-1mm}\textbf{FDHT-LSTM}\hspace{-1mm}  & \multirow{2}{*}{N/A}  & \textbf{8,324}  & \multirow{2}{*}{\textbf{88.2}}      \\
\textbf{(Ours)} & & (\textbf{7,117$\times$}) & \\
\hline
\end{tabular}

\label{tab:directly_ytc}

\end{table}

Besides, we also compare the performance of FDHT-LSTM with several other tensor decomposition-free models for video recognition, and the results are summarized in Table \ref{tab:directly_ytc2}. It is seen that our FDHT-LSTM can outperform the state-of-the-art work \cite{li2018face} with 3.6\% higher accuracy and much smaller model size.

\begin{table}[t]
\normalsize
\caption{Experimental comparison among FDHT-LSTM and other non-tensor decomposition models, e.g., DML-PV \cite{cheng2017duplex}, VGGFACE + RRNN \cite{li2018face} and VGG16-GCR \cite{liu2019group}, on Youtube celebrities face dataset.}
\def\arraystretch{1.05}
\centering
\begin{tabular}{l|c|c}  
\hline
 \multicolumn{1}{c|}{\multirow{2}{*}{\textbf{Model}}}  & \multirow{2}{*}{\makecell{\textbf{Number of}\\\textbf{Parameters}}}& \textbf{Top-1} \\
 & & \textbf{Acc.} (\%)\\
\hline
\hline
DML-PV  & 220K &  82.8 \\
\hline
VGGFACE + RRNN  & $\geq$42M  & 84.6 \\
\hline
VGG16-GCR  & 138M & 82.9  \\
\hline
HT-LSTM (Ours) & 263K & 88.1 \\
\hline
FDHT-LSTM (Ours)  & \textbf{8,324} & \textbf{88.2} \\
\hline
\end{tabular}

\label{tab:directly_ytc2}
\end{table}

\begin{table}[t]
\normalsize
\caption{Accuracy comparison with various $R$ on LSTM-UCF11 and LSTM-Youtube.}
\centering
\begin{tabular}{@{}ccc@{}}
\toprule
\textbf{Models}               &\textbf{$R$}  & \textbf{Top-1 Acc.(\%)} \\ \hline
\hline
\multirow{3}{*}{LSTM-UCF11}   & 2 & 81.4 \\
  & \textbf{12} & \textbf{87.5} \\
  & 32 & 79.3             \\ \midrule
\multirow{3}{*}{LSTM-Youtube} & 2 & 78.9 \\
 & \textbf{11} & \textbf{88.2}\\
  & 32 & 77.4 \\ \bottomrule
\end{tabular}
\label{tab:accvsR}
\end{table}

\textbf{Rank Selection.} As analyzed before, for compressing large LSTM models, HT decomposition can improve accuracy with very high compression ratio because it can effectively reduces the structural redundancy. Therefore, for high rank setting such as $R=32$, the accuracy is not as good as low rank setting because considerable unnecessary redundancy still exists and it is not friendly for training. On the other hand, for extremely low rank setting, e.g., $R=2$, the model size is too small and thus the model capacity will be hurt. Therefore, as shown in Table \ref{tab:accvsR}, in order to achieve good balance between compression rate and model accuracy, $R=12$ and $R=11$ are set for LSTM-UCF11 and LSTM-Youtube, respectively.

\subsection{Performance of FDHT-LSTM Hardware}

\subsubsection{Configuration of Experiment and Design Example}
A bit-accurate cycle-accurate simulator is developed to model the high-level behavior of the proposed FDHT-LSTM architecture.
Then, we build a Verilog-based RTL model and verify the functional validation. This model is synthesized with CMOS 28nm library. Table~\ref{tab:design-configuration} shows the detailed configuration information of the design example based on the proposed FDHT-LSTM architecture. 
Here the overall hardware consists of $16$ PEs, and each PE is equipped with 16 16-bit multipliers and 16 24-bit accumulators, hence 256 MAC operations are performed in each clock cycle. The working memory is composed of $14$ banks of SRAM, and each SRAM bank has the width of 256 bits and the depth of 2048, thereby leading to 875KB budgeted capacity that is sufficient for storing the activation and intermediate result in most applications. Since the working SRAM contains two copies to serve as a ping-pong buffer, the total size of the working SRAM is $875$KB $\times 2 = 1.70$MB. For the weight SRAM, it has 16-bit width with depth of 8808 to store the weight parameters of the compressed FDHT-LSTM models. Thanks to the ultra-strong compression capability of FDHT technique, the budgeted 17.2KB weight SRAM can support the storage of many large-scale LSTM models.


\begin{table}[t]
\caption{Architectural configuration for design example.}
\centering
\normalsize
\def\arraystretch{1.1}
\label{tab:design-configuration}
\begin{tabular}{cccl}
\hline
\multicolumn{2}{c|}{\textbf{System Parameter}}                                    & \multicolumn{2}{c}{\textbf{Value}} \\ \hline
\hline
\multicolumn{2}{c|}{PE}                                                           & \multicolumn{2}{c}{16}             \\ \hline
\multicolumn{2}{c|}{Quantization}                                                 & \multicolumn{2}{c}{16-bit}         \\ \hline
\multicolumn{4}{c}{\textbf{Memory Parameter}}                                                                          \\ \hline
\multicolumn{1}{c|}{\multirow{4}{*}{Working Memory}} & \multicolumn{1}{c|}{Width (W)}     & \multicolumn{2}{c}{256-bit}        \\ \cline{2-4} 
\multicolumn{1}{c|}{}                                & \multicolumn{1}{c|}{Depth (M)}     & \multicolumn{2}{c}{2048}           \\ \cline{2-4} 
\multicolumn{1}{c|}{}                                & \multicolumn{1}{c|}{\# of Banks (G)}     & \multicolumn{2}{c}{14}             \\ \cline{2-4} 
\multicolumn{1}{c|}{}                                & \multicolumn{1}{c|}{Total Size}  & \multicolumn{2}{c}{1.7MB}          \\ \hline
\multicolumn{1}{c|}{\multirow{3}{*}{Weight Memory}}  & \multicolumn{1}{c|}{Width} & \multicolumn{2}{c}{16-bit}         \\ \cline{2-4} 
\multicolumn{1}{c|}{}                                & \multicolumn{1}{c|}{Depth} & \multicolumn{2}{c}{8808}           \\ \cline{2-4} 
\multicolumn{1}{c|}{}                                & \multicolumn{1}{c|}{Size}  & \multicolumn{2}{c}{17.2KB}         \\ \hline
\multicolumn{4}{c}{\textbf{PE Parameter}}                                                                              \\ \hline
\multicolumn{2}{c|}{Amount of Multiplier}                                         & \multicolumn{2}{c}{16}             \\ \hline
\multicolumn{2}{c|}{Amount of Accumulator}                                        & \multicolumn{2}{c}{16}             \\ \hline
\end{tabular}
\end{table}


\subsubsection{Comparison with GPU}

Though the decomposed model reduces the weight size significantly and enjoys the parallel processing, the computation flow of FDHT-LSTM contains complicated matrix transformation, which is not naturally supported by GPU in an efficient way. We implement the decomposed LSTM-Youtube and LSTM-UCF11 on NVIDIA RTX A6000, and compare the inference speed with our FDHT hardware accelerator. The result shows that FDHT-LSTM hardware achieves \textbf{2.30$\times$} and \textbf{1.85$\times$} inference speedup than GPU for executing LSTM-Youtube and LSTM-UCF11, respectively.

\subsubsection{Comparison with EIE and TIE}
In this subsection, we compare the proposed architecture with two prior compressed model-oriented hardware accelerators EIE \cite{han_eie_2016} and TIE \cite{deng_tie_2019}. Table~\ref{tab:com-EIE-TIE} summarizes the implementation results for the three listed ASIC designs. Table \ref{tab:FDHTvsEIE_acc} shows the compression performance of FDHT-LSTM, EIE and TIE, where top-1 accuracy is the accuracy that the recognition result with the highest probability is exactly the expected answer. It is seen that our FDHT-LSTM outperforms EIE and TIE with respect to compression rate and accuracy performance on both UCF11 and Youtube datasets.
\begin{table}[t]
\caption{Compression performance of EIE, TIE and our FDHT-LSTM on UCF11 and Youtube dataset. Top-1 accuracy is the accuracy that the recognition result with the highest probability is exactly the expected answer.}
\centering
\normalsize
\begin{tabular}{@{}cccc@{}}
\toprule
& \textbf{Design} & \textbf{Compres. Rate} & \textbf{Top-1 Acc.(\%)} \\ 
\hline
\hline
\multirow{3}{*}{UCF11} & EIE & 117$\times$ & 82.9 \\
& TIE & 4954$\times$ & 79.6 \\
& \textbf{\begin{tabular}[c]{@{}c@{}}FDHT-LSTM\\ (Ours)\end{tabular}} & \textbf{6726$\times$} & \textbf{87.5} \\ \midrule
\multirow{3}{*}{Youtube} & EIE & 111$\times$ & 84.2 \\
& TIE & 4608$\times$ & 78.8 \\ & \textbf{\begin{tabular}[c]{@{}c@{}}FDHT-LSTM\\ (Ours)\end{tabular}} & \textbf{7117$\times$} & \textbf{88.2} \\ \bottomrule
\end{tabular}
\label{tab:FDHTvsEIE_acc}
\end{table}



\begin{figure*}
    \centering
    \includegraphics[width=1.7\columnwidth]{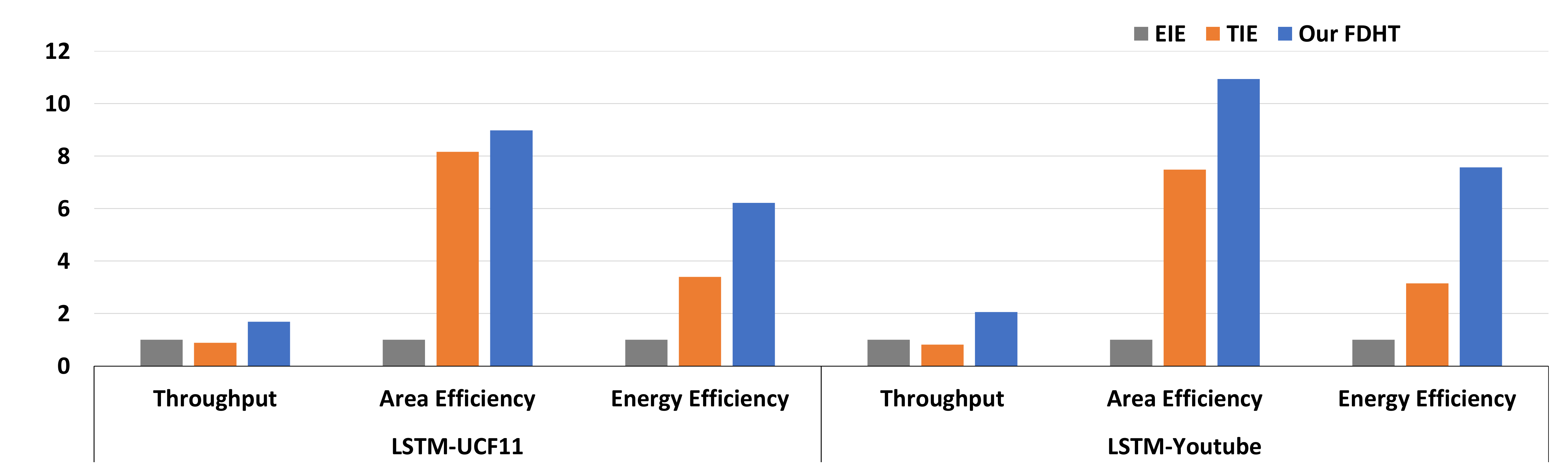}
    \caption{Hardware performance comparison among FDHT-LSTM, TIE and EIE.}
    \label{fig:com-EIE-TIE}
\end{figure*}

\textbf{Comparison with EIE}. EIE is a sparse model-oriented hardware architecture. Consider EIE is implemented with a different technology node, as reported in Table \ref{tab:com-EIE-TIE}, the performance metrics of EIE are projected to 28nm technology for fair comparison. Here the projection follows the scaling rule adopted in \cite{han_eie_2016}: linear, quadratic and constant scaling for clock frequency, silicon area and power consumption, respectively. In addition, considering the different budgeted arithmetic and memory resource,
Figure~\ref{fig:com-EIE-TIE} compares our proposed FDHT-LSTM with EIE with respect to different hardware performance metrics. It is seen that the FDHT-LSTM achieves $1.69\times$, $8.99\times$ and $6.22\times$ increase in throughput, area efficiency and energy efficiency on UCF11, and $2.06\times$, $10.94\times$ and $7.58\times$ increase in throughput, area efficiency and energy efficiency on Youtube respectively.

\textbf{Comparison with TIE.} TIE is the state-of-the-art tensor train (TT) decomposed model-oriented hardware accelerator, which is the most related work to our FDHT-LSTM architecture. Figure~\ref{fig:com-EIE-TIE} shows the performance comparison between the two designs in terms of processing throughput, area efficiency and energy efficiency on two LSTM workloads \footnote{Because TIE is based on the TT decomposition that only compresses the input-to-hidden layer of the original LSTM,  the reported performance of TIE is evaluated on that single layer.}. It is seen that for executing LSTM-Youtube workload, our proposed FDHT-LSTM architecture achieves $2.5\times$, $1.46\times$ and $2.41\times$ increase in throughput, area efficiency and $2.41\times$ energy efficiency, respectively. For LSTM-UCF workload, our FDHT-LSTM design also outperforms TIE with $1.9\times$ higher throughput, $1.83\times $ higher energy efficiency and comparable area efficiency.

\begin{table}[t]
\caption{Comparison on hardware performance.}
\normalsize
\centering
\setlength\tabcolsep{5.2pt}
\begin{tabular}{c|c|c|c}
\hline
\textbf{Design}              & \textbf{EIE \cite{han_eie_2016}}                                                                               & \textbf{TIE \cite{deng_tie_2019}} & \textbf{Our FDHT} \\ \hline
\hline

\textbf{Compression}  & \multirow{2}{*}{Pruning} & Tensor Train & Hierarchical  \\
\textbf{Approach} & & (TT) &  Tucker (HT)\\

\hline
\textbf{CMOS} & 28nm & \multirow{2}{*}{28nm}  & \multirow{2}{*}{28nm}  \\
\textbf{Technology} & (projected) & & \\
\hline
\textbf{Frequency} & \multirow{2}{*}{1280}   & \multirow{2}{*}{1000}  & \multirow{2}{*}{1000}  \\
\textbf{(MHz)} & & & \\
\hline
\textbf{Quantization} & 4-bit index &  \multirow{2}{*}{16-bit} & \multirow{2}{*}{16-bit}    \\
\textbf{Scheme} & 16-bit weight & & \\
\hline
\textbf{Area (mm)}           & 15.7                                                                                       & 1.74         & 2.96          \\ \hline
\textbf{Power (mW)}          & 590                                                                                        & 154.8        & 160.4         \\ \hline
\end{tabular}
\label{tab:com-EIE-TIE}
\end{table}


\subsubsection{Flexibility}

\begin{figure}[t]
    \centering
    \includegraphics[width=\columnwidth]{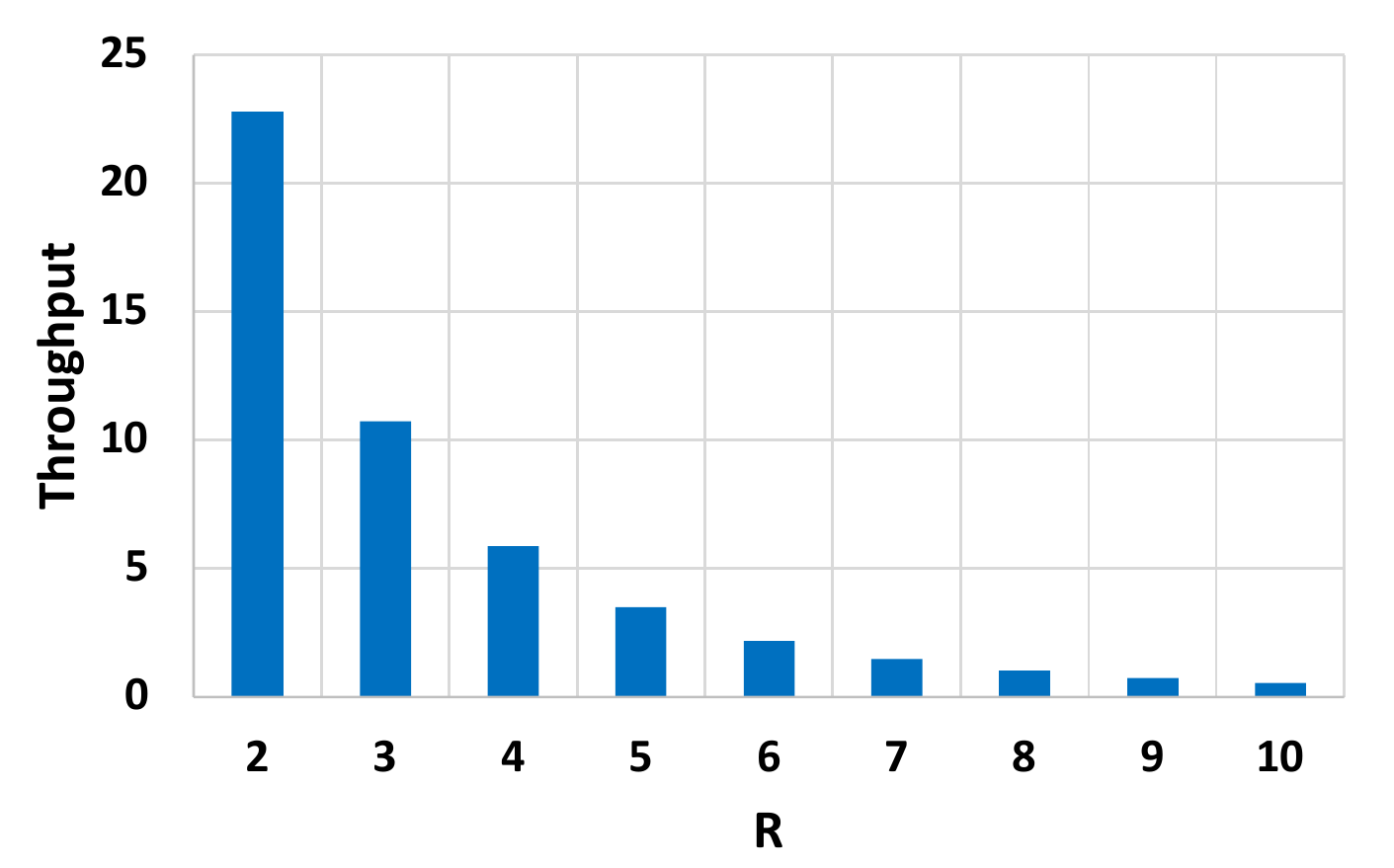}
    \caption{Flexibility of FDHT with respect to different decomposition ranks.}
    \label{fig:throughput-ranks}
\end{figure}

The proposed FDHT-LSTM architecture can provide high flexibility to support different compressed LSTM models with different topology settings. Considering the rank $R$ is the most important parameters that directly determines the storage requirement and computational costs of the underlying models, we study the flexibility of the proposed hardware architecture with respect to different $R$'s. As reported in Figure \ref{fig:throughput-ranks}, FDHT-LSTM hardware demonstrates strong flexibility.

\section{Related Work}

Customized hardware accelerators for deep neural networks have been extensively studied in recent years. To accelerate the execution of convolutional neural networks (CNNs) in a real-time and energy-efficient way, a series of CNN hardware architecture have been proposed in \cite{chen2016eyeriss} \cite{zhang2015optimizing} \cite{alwani2016fused}. In particular, fully leveraging the sparsity is a very important optimization strategy towards high-performance CNN hardware accelerators \cite{parashar2017scnn}\cite{aimar2018nullhop}\cite{gondimalla2019sparten}.

Different from computation-bounded CNNs, LSTMs are inherently memory-bounded and require very large model size because of its component linear layer. Therefore, the focus of efficient LSTM hardware design is to efficiently integrate upper-level model compression algorithm to the lower-level architecture optimization. Motivated by this design philosophy, several compressed model-oriented LSTM hardware have been developed in \cite{wang2018c}\cite{han2017ese}\cite{han_eie_2016}\cite{deng_tie_2019}. To be specific, EIE\cite{han_eie_2016} and ESE \cite{han2017ese} are built on pruned model and the focus of their architectural optimization is to properly leverage sparsity opportunities. C-LSTM \cite{wang2018c} aims to use FFT-based compression compression to accelerate LSTM, and hence its underlying architecture mainly consists of optimized FFT modules. TIE is the first tensor decomposition-based DNN hardware architecture, which is the most closed work to our design. By leveraging the tensor train decomposition, TIE \cite{deng_tie_2019} can achieve competitive hardware performance with good resource utilization.

\section{Conclusion}
\label{sec:conclu}

In this paper, we propose algorithm and hardware co-design for FDHT-LSTM, a ultra-compact high-performance compressed LSTM network. By fully decomposing the entire LSTM models via hierarchical Tucker decomposition, the entire LSTM model size can be significantly reduced. Meanwhile, an energy-efficient customized hardware architecture with delicate design of memory access scheme is developed. The proposed algorithm and hardware are empirically evaluated on different video recognition datasets and LSTM workloads. The experimental results show that our proposed FDHT-LSTM networks and the corresponding hardware accelerator can achieve very high model performance and hardware performance as compared to the state-of-the-art designs.


%

\ifCLASSOPTIONcompsoc
  \section*{Acknowledgments}
\else
  \section*{Acknowledgment}
\fi
This work was partially supported by National Science Foundation under Grant CCF-1955909.
\ifCLASSOPTIONcaptionsoff
  \newpage
\fi


{\small
\bibliographystyle{ieee_fullname}
\bibliography{ref}

\begin{thebibliography}{10}\itemsep=-1pt

\bibitem{aimar2018nullhop}
Alessandro Aimar, Hesham Mostafa, Enrico Calabrese, Antonio Rios-Navarro,
  Ricardo Tapiador-Morales, Iulia-Alexandra Lungu, Moritz~B Milde, Federico
  Corradi, Alejandro Linares-Barranco, Shih-Chii Liu, et~al.
\newblock Nullhop: A flexible convolutional neural network accelerator based on
  sparse representations of feature maps.
\newblock {\em IEEE transactions on neural networks and learning systems},
  30(3):644--656, 2018.

\bibitem{alwani2016fused}
Manoj Alwani, Han Chen, Michael Ferdman, and Peter Milder.
\newblock Fused-layer cnn accelerators.
\newblock In {\em 2016 49th Annual IEEE/ACM International Symposium on
  Microarchitecture (MICRO)}, pages 1--12. IEEE, 2016.

\bibitem{azari_energy-efficient_2019}
Elham Azari and Sarma Vrudhula.
\newblock An {Energy}-{Efficient} {Reconfigurable} {LSTM} {Accelerator} for
  {Natural} {Language} {Processing}.
\newblock In {\em 2019 {IEEE} {International} {Conference} on {Big} {Data}
  ({Big} {Data})}, pages 4450--4459, Los Angeles, CA, USA, Dec. 2019. IEEE.

\bibitem{chen2016eyeriss}
Yu-Hsin Chen, Joel Emer, and Vivienne Sze.
\newblock Eyeriss: A spatial architecture for energy-efficient dataflow for
  convolutional neural networks.
\newblock {\em ACM SIGARCH Computer Architecture News}, 44(3):367--379, 2016.

\bibitem{cheng2017duplex}
Gong Cheng, Peicheng Zhou, and Junwei Han.
\newblock Duplex metric learning for image set classification.
\newblock {\em IEEE Transactions on Image Processing}, 27(1):281--292, 2017.

\bibitem{deng_tie_2019}
Chunhua Deng, Fangxuan Sun, Xuehai Qian, Jun Lin, Zhongfeng Wang, and Bo Yuan.
\newblock {TIE}: energy-efficient tensor train-based inference engine for deep
  neural network.
\newblock In {\em Proceedings of the 46th {International} {Symposium} on
  {Computer} {Architecture}}, pages 264--278, Phoenix Arizona, June 2019. ACM.

\bibitem{donahue_long-term_nodate}
Jeffrey Donahue, Lisa~Anne Hendricks, Sergio Guadarrama, Marcus Rohrbach,
  Subhashini Venugopalan, Kate Saenko, and Trevor Darrell.
\newblock Long-{Term} {Recurrent} {Convolutional} {Networks} for {Visual}
  {Recognition} and {Description}.
\newblock page~10.

\bibitem{gondimalla2019sparten}
Ashish Gondimalla, Noah Chesnut, Mithuna Thottethodi, and TN Vijaykumar.
\newblock Sparten: A sparse tensor accelerator for convolutional neural
  networks.
\newblock In {\em Proceedings of the 52nd Annual IEEE/ACM International
  Symposium on Microarchitecture}, pages 151--165, 2019.

\bibitem{graves_hybrid_2013}
Alex Graves, Navdeep Jaitly, and Abdel-rahman Mohamed.
\newblock Hybrid speech recognition with {Deep} {Bidirectional} {LSTM}.
\newblock In {\em 2013 {IEEE} {Workshop} on {Automatic} {Speech} {Recognition}
  and {Understanding}}, pages 273--278, Olomouc, Czech Republic, Dec. 2013.
  IEEE.

\bibitem{hackbusch2009new}
Wolfgang Hackbusch and Stefan K{\"u}hn.
\newblock A new scheme for the tensor representation.
\newblock {\em Journal of Fourier Analysis and Applications}, 15(5):706--722,
  2009.

\bibitem{han2017ese}
Song Han, Junlong Kang, Huizi Mao, Yiming Hu, Xin Li, Yubin Li, Dongliang Xie,
  Hong Luo, Song Yao, Yu Wang, et~al.
\newblock Ese: Efficient speech recognition engine with sparse lstm on fpga.
\newblock In {\em Proceedings of the 2017 ACM/SIGDA International Symposium on
  Field-Programmable Gate Arrays}, pages 75--84, 2017.

\bibitem{han_eie_2016}
Song Han, Xingyu Liu, Huizi Mao, Jing Pu, Ardavan Pedram, Mark~A. Horowitz, and
  William~J. Dally.
\newblock {EIE}: {Efficient} {Inference} {Engine} on {Compressed} {Deep}
  {Neural} {Network}.
\newblock In {\em 2016 {ACM}/{IEEE} 43rd {Annual} {International} {Symposium}
  on {Computer} {Architecture} ({ISCA})}, pages 243--254, Seoul, South Korea,
  June 2016. IEEE.

\bibitem{kim2008face}
Minyoung Kim, Sanjiv Kumar, Vladimir Pavlovic, and Henry Rowley.
\newblock Face tracking and recognition with visual constraints in real-world
  videos.
\newblock In {\em Proceedings of the IEEE Conference on Computer Vision and
  Pattern Recognition}, pages 1--8. IEEE, 2008.

\bibitem{li2018face}
Yang Li, Wenming Zheng, Zhen Cui, and Tong Zhang.
\newblock Face recognition based on recurrent regression neural network.
\newblock {\em Neurocomputing}, 297:50--58, 2018.

\bibitem{liu2019group}
Bo Liu, Liping Jing, Jia Li, Jian Yu, Alex Gittens, and Michael~W Mahoney.
\newblock Group collaborative representation for image set classification.
\newblock {\em International Journal of Computer Vision}, 127(2):181--206,
  2019.

\bibitem{liu2009recognizing}
Jingen Liu, Jiebo Luo, and Mubarak Shah.
\newblock Recognizing realistic actions from videos “in the wild”.
\newblock In {\em Proceedings of the IEEE Conference on Computer Vision and
  Pattern Recognition}, pages 1996--2003. IEEE, 2009.

\bibitem{pan2019compressing}
Yu Pan, Jing Xu, Maolin Wang, Jinmian Ye, Fei Wang, Kun Bai, and Zenglin Xu.
\newblock Compressing recurrent neural networks with tensor ring for action
  recognition.
\newblock In {\em Proceedings of the AAAI Conference on Artificial
  Intelligence}, volume~33, pages 4683--4690, 2019.

\bibitem{parashar2017scnn}
Angshuman Parashar, Minsoo Rhu, Anurag Mukkara, Antonio Puglielli, Rangharajan
  Venkatesan, Brucek Khailany, Joel Emer, Stephen~W Keckler, and William~J
  Dally.
\newblock Scnn: An accelerator for compressed-sparse convolutional neural
  networks.
\newblock {\em ACM SIGARCH Computer Architecture News}, 45(2):27--40, 2017.

\bibitem{wang2018c}
Shuo Wang, Zhe Li, Caiwen Ding, Bo Yuan, Qinru Qiu, Yanzhi Wang, and Yun Liang.
\newblock C-lstm: Enabling efficient lstm using structured compression
  techniques on fpgas.
\newblock In {\em Proceedings of the 2018 ACM/SIGDA International Symposium on
  Field-Programmable Gate Arrays}, pages 11--20, 2018.

\bibitem{yang2017tensor}
Yinchong Yang, Denis Krompass, and Volker Tresp.
\newblock Tensor-train recurrent neural networks for video classification.
\newblock In {\em International Conference on Machine Learning}, pages
  3891--3900. JMLR. org, 2017.

\bibitem{ye2018learning}
Jinmian Ye, Linnan Wang, Guangxi Li, Di Chen, Shandian Zhe, Xinqi Chu, and
  Zenglin Xu.
\newblock Learning compact recurrent neural networks with block-term tensor
  decomposition.
\newblock In {\em Proceedings of the IEEE Conference on Computer Vision and
  Pattern Recognition}, pages 9378--9387, 2018.

\bibitem{yin2021towards}
Miao Yin, Siyu Liao, Xiao-Yang Liu, Xiaodong Wang, and Bo Yuan.
\newblock Towards extremely compact rnns for video recognition with fully
  decomposed hierarchical tucker structure.
\newblock In {\em Proceedings of the IEEE/CVF Conference on Computer Vision and
  Pattern Recognition}, pages 12085--12094, 2021.

\bibitem{yu2017long}
Rose Yu, Stephan Zheng, Anima Anandkumar, and Yisong Yue.
\newblock Long-term forecasting using higher order tensor rnns.
\newblock {\em arXiv preprint arXiv:1711.00073}, 2017.

\bibitem{zhang2015optimizing}
Chen Zhang, Peng Li, Guangyu Sun, Yijin Guan, Bingjun Xiao, and Jason Cong.
\newblock Optimizing fpga-based accelerator design for deep convolutional
  neural networks.
\newblock In {\em Proceedings of the 2015 ACM/SIGDA international symposium on
  field-programmable gate arrays}, pages 161--170, 2015.

\end{thebibliography}
}
%



%

\begin{IEEEbiography}[{\includegraphics[width=1in,height=1.25in,clip,keepaspectratio]{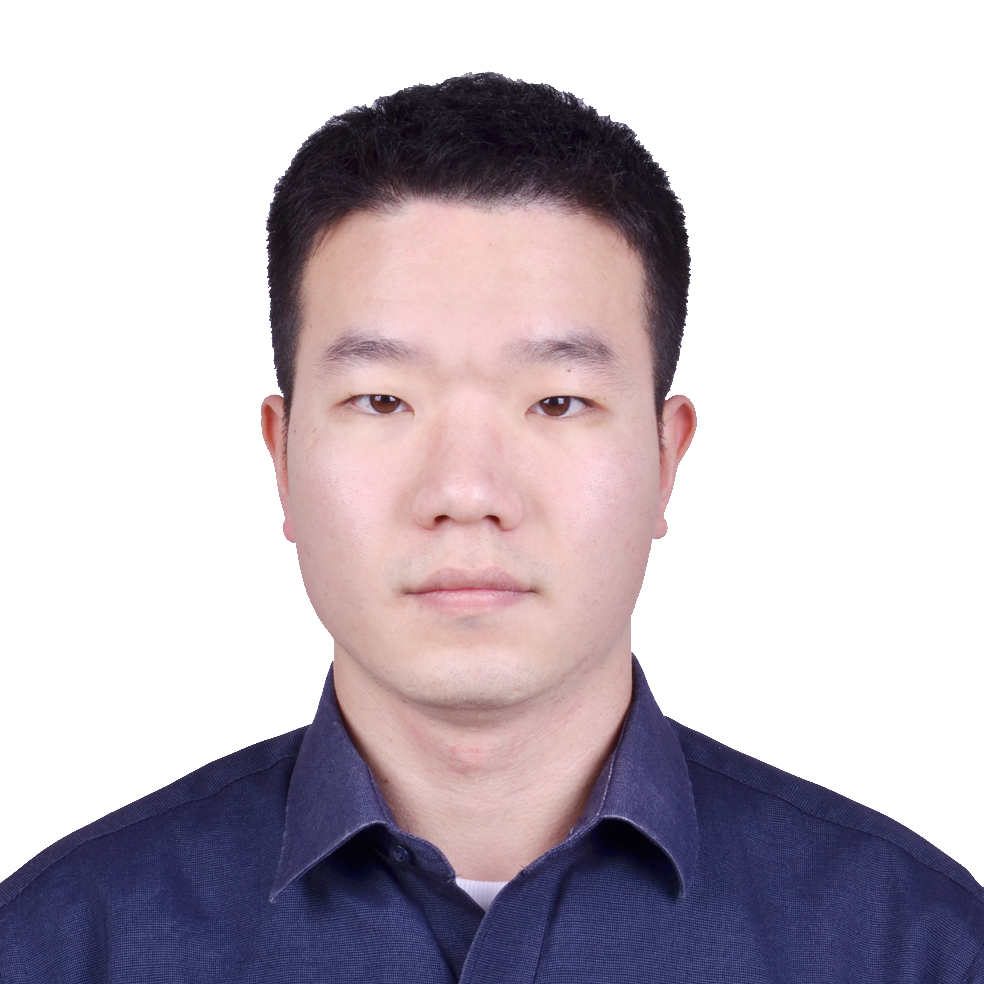}}]{Yu Gong}
received the master’s degree in Shanghai Jiao Tong University, and bachelor's degree in Wuhan University of Technology. 
Currently, he is a Ph.D. student in Electrical and Computer Engineering at Rutgers University. His research interests include high-performance architecture for AI and hardware-software co-design.
\end{IEEEbiography}

\begin{IEEEbiography}[{\includegraphics[width=1in,height=1.25in,clip,keepaspectratio]{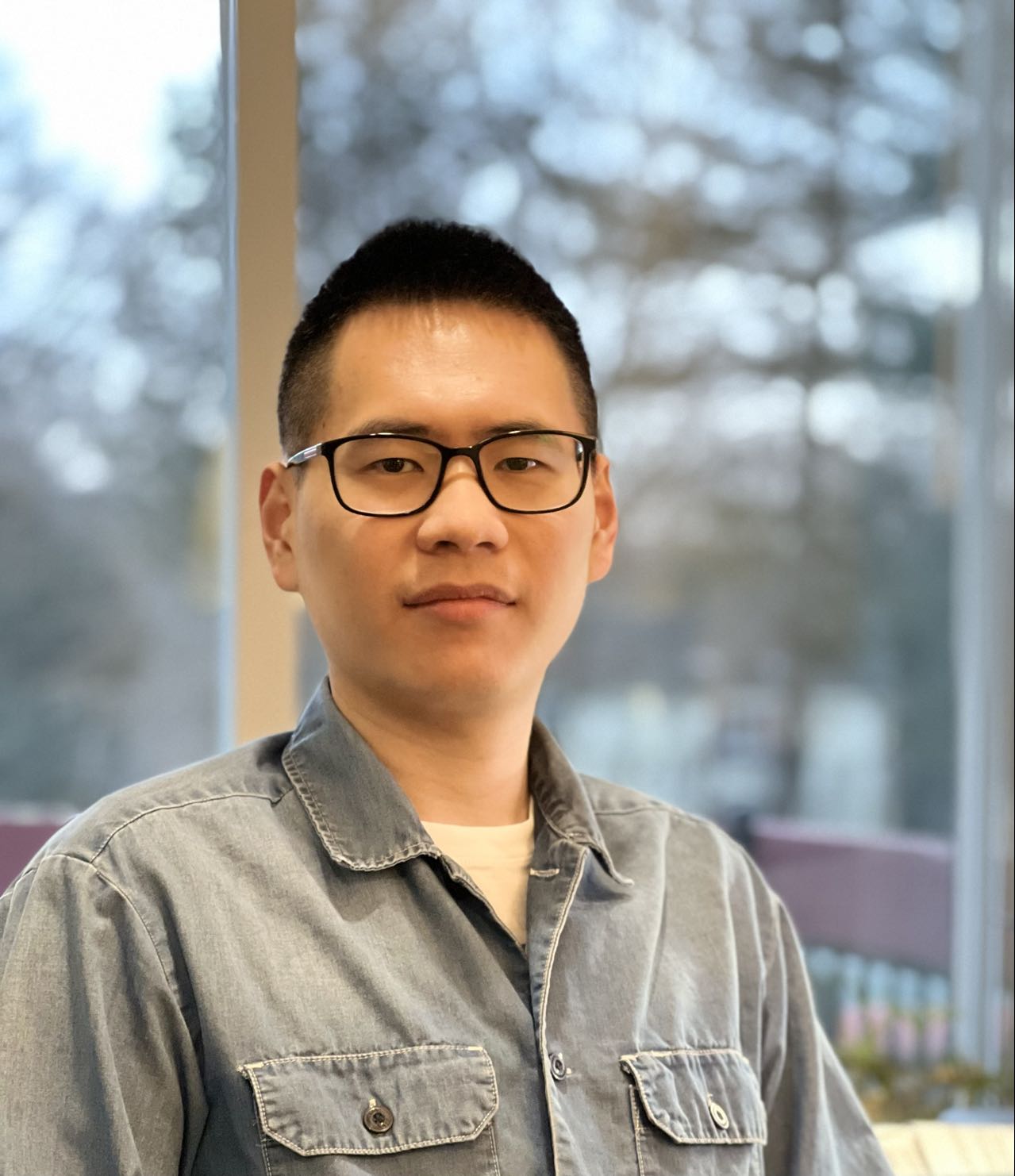}}]{Miao Yin}
is a Ph.D. student in Electrical and Computer Engineering at Rutgers, The State University of New Jersey. His research is focused on algorithm-hardware co-designed energy-efficient AI system (e.g., on-device
image and action recognition) based on higher-order tensor decomposition and advanced optimization. He serves as a PC member of the IEEE/CVF Conference on Computer Vision and Pattern Recognition.
\end{IEEEbiography}

\begin{IEEEbiography}[{\includegraphics[width=1in,height=1.25in,clip,keepaspectratio]{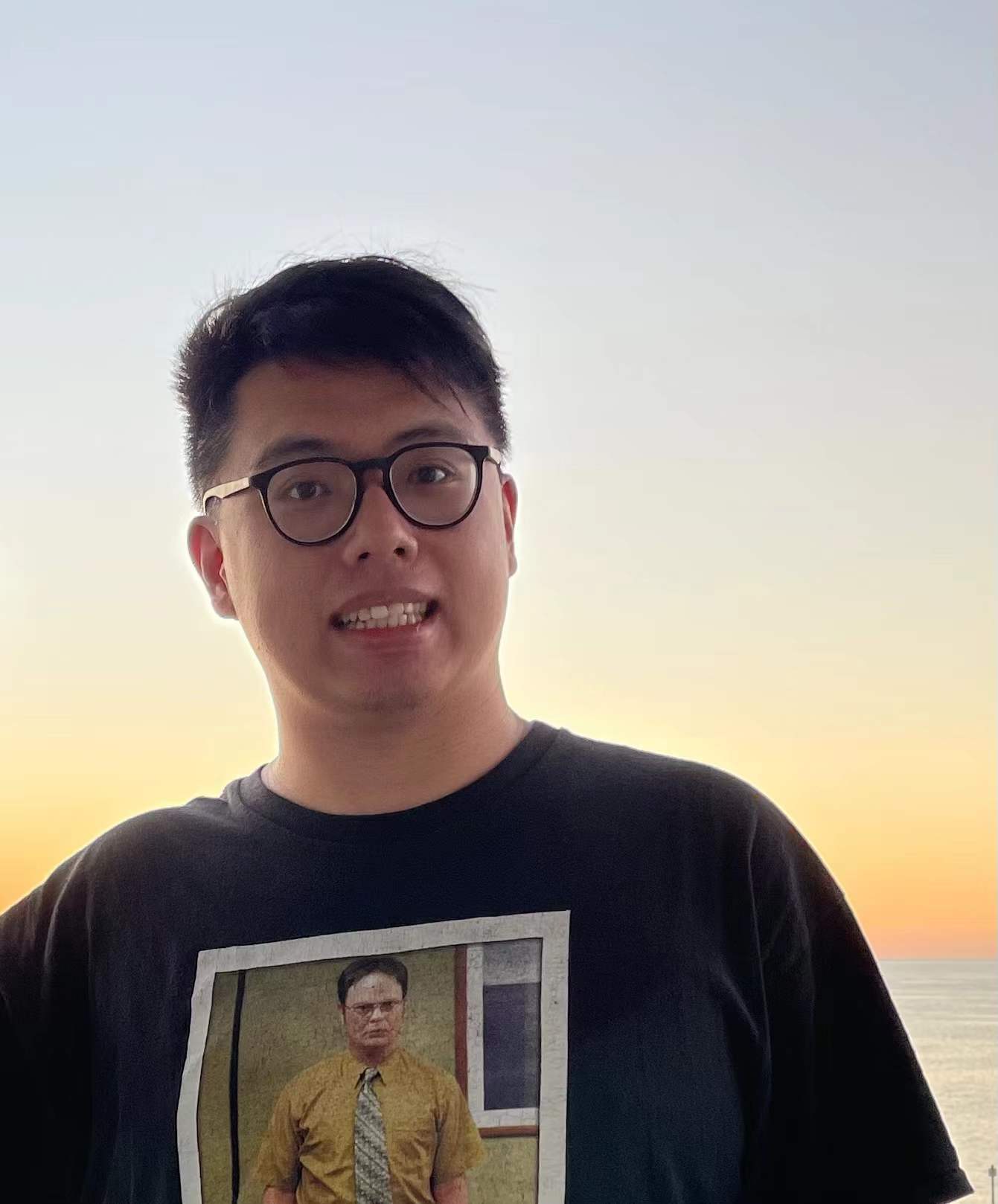}}]{Lingyi Huang}
received the master's degree in electrical engineering from New York University and the bachelor's degree in electrical engineering from Xidian University. He is currently pursuing the PhD degree in computer engineering at Rutgers University.
\end{IEEEbiography}

\begin{IEEEbiography}[{\includegraphics[width=1in,height=1.25in,clip,keepaspectratio]{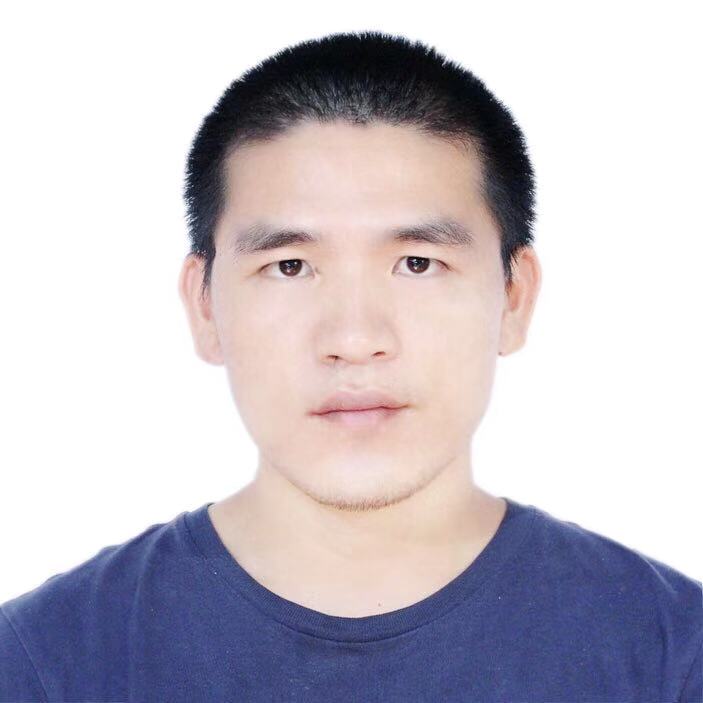}}]{Chunhua Deng}
is a senior staff design engineer at ScaleFlux Inc. He received his Ph.D. degree from the Department of Electrical and Computer Engineering at Rutgers University. He received his bachelor's degree and master's degree from China University of Petroleum, Beijing Institute of Technology, respectively. His research interests include machine learning, computer architecture, and VLSI design.
\end{IEEEbiography}

\begin{IEEEbiography}[{\includegraphics[width=1in,height=1.25in,clip,keepaspectratio]{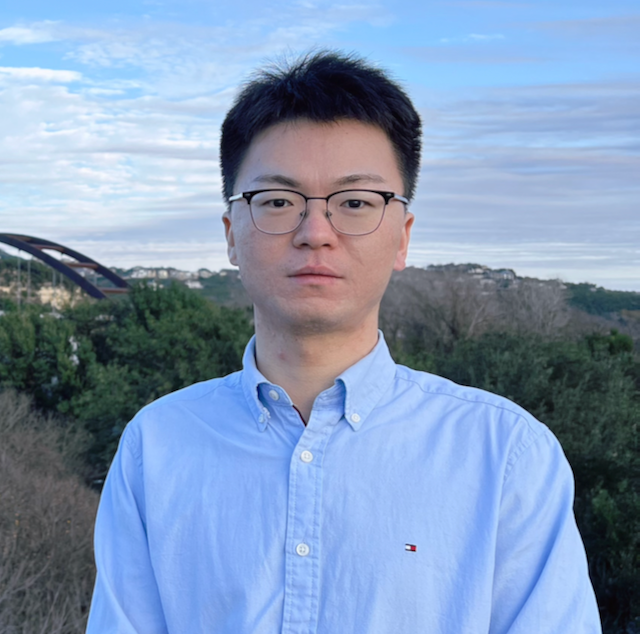}}]{Yang Sui}
is a Ph.D. student in Electrical and Computer Engineering at Rutgers, The State University of New Jersey. His research is focusing on algorithm-hardware co-designed efficient AI with model compression based on pruning, low-rank decomposition, and advanced algorithms.
\end{IEEEbiography}

\begin{IEEEbiography}[{\includegraphics[width=1in,height=1.25in,clip,keepaspectratio]{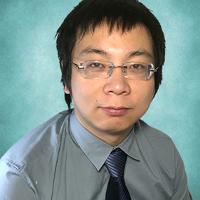}}]{Bo Yuan}
received the bachelor’s and master’s degrees from Nanjing University, Nanjing, China, in 2007 and 2010, respectively, and the PhD degree from the Department of Electrical and Computer Engineering, University of Minnesota, Twin Cities, Minnesota, in 2015. His research interests include algorithm and hardware co-design and implementation for machine learning and signal processing systems, error-resilient low-cost computing techniques for embedded and IoT systems and machine learning for domain-specific applications. He is the recipient of Global Research Competition Finalist Award in Broadcom Corporation and doctoral dissertation fellowship with the University of Minnesota. He serves as technical committee track chair and technical committee member for several IEEE/ACM conferences. He is the associated editor of the Springer Journal of Signal Processing System.
\end{IEEEbiography}




\vfill


\end{document}